\documentclass{article} 
\PassOptionsToPackage{table}{xcolor}
\usepackage{iclr2027_conference,times}

\usepackage{amsmath,amsfonts,bm}

\def\eqref#1{equation~\ref{#1}}

\def\1{\bm{1}}

\DeclareMathAlphabet{\mathsfit}{\encodingdefault}{\sfdefault}{m}{sl}
\SetMathAlphabet{\mathsfit}{bold}{\encodingdefault}{\sfdefault}{bx}{n}

\DeclareMathOperator*{\argmax}{arg\,max}

\usepackage{graphicx}

\usepackage{wrapfig}
\usepackage{amsmath}
\usepackage{amssymb}
\usepackage{booktabs}
\usepackage{multirow}
\usepackage{xcolor}
\definecolor{bestcell}{HTML}{7AB0D8}
\definecolor{secondcell}{HTML}{D9EAF7}
\newcommand{\best}[1]{\cellcolor{bestcell}{#1}}
\newcommand{\second}[1]{\cellcolor{secondcell}{#1}}
\usepackage{graphicx}
\usepackage{subcaption}

\usepackage[most]{tcolorbox}
\usepackage{listings}
\usepackage{amssymb}

\tcbset{aibox/.style={
    breakable,
    colback=black!8,
    boxsep=3pt,
    left=4pt,
    right=4pt,
    before skip=6pt,
    after skip=6pt,
}}
\lstnewenvironment{promptcode}{%
  \lstset{
    basicstyle=\ttfamily\footnotesize,
    breaklines=true,
    breakindent=0pt,
    breakautoindent=false,
    breakatwhitespace=false,
    columns=fullflexible,
    keepspaces=true,
    showstringspaces=false,
    frame=none,
    aboveskip=0pt,
    belowskip=0pt,
    xleftmargin=0pt,
    xrightmargin=0pt,
  }%
}{}
\newcommand{\harnessname}{E-LEDGER}
\newcommand{\evolutionname}{WorldAbduct}
\newtcolorbox{AIbox}[2][]{aibox,title={#2},#1}
\usepackage{hyperref}
\usepackage{url}

\title{Safe, Persistent, and Evolving Agent Harness for Understanding Partially Observable Worlds}

\author{%
\normalfont
\makebox[\dimexpr\textwidth-2\tabcolsep\relax][c]{%
\begin{tabular}{@{}c@{}}
Yisen Gao$^{1}$, Yue Guo$^{2}$, Qing Zong$^{1}$,
Yiwen Guo$^{3}$\thanks{Yiwen Guo is the corresponding author.}, Yangqiu Song$^{1}$ \\[3pt]
$^{1}$The Hong Kong University of Science and Technology \\
$^{2}$LIGHTSPEED\\
$^{3}$Independent Researcher \\[3pt]
\texttt{\{ygaodi\}@cse.ust.hk}
\end{tabular}}%
}

\iclrfinalcopy 
\begin{document}

\maketitle
\lhead{Preprint.}

\begin{abstract}
Large language model agents can invoke tools fluently, but enterprise
workflows demand more than selecting the right tools: actions must strictly
comply with organizational policies, tool feedback often conceals hidden side
effects under partial observability, and long-horizon tasks require persistent
state tracking across multiple records.  To address these challenges, we
introduce E-Ledger, a multi-agent harness for safe and persistent execution.
E-Ledger employs a code approval layer that checks every proposed action
against policy before execution, and maintains a world ledger of verified hidden rules
alongside evidence-backed dynamic state.  Because hidden rules are typically
unknown a priori, we further propose WorldAbduct, an abductive, world-model-driven
harness evolution framework.  WorldAbduct diagnoses execution trajectories
across four complementary views (state consistency, world-observation gap,
policy-gate correctness, and goal judgment) to hypothesize latent rules,
and verifies them through targeted abductive interactions before integrating
them into the ledger.  On the enterprise benchmark World of Workflows,
E-Ledger with WorldAbduct improves safe task completion across four LLM
backbones, outperforming the strongest evolution baseline by 5--15
percentage points. Experiments in ScienceWorld and DiscoveryWorld further
show that abductive harness evolution carries over to scientific
environments. Our code is available at https://github.com/HKUST-KnowComp/E-LEDGER-WorldAbduct.
\end{abstract}
\section{Introduction}

Large language model (LLM) agents increasingly complete multi-step tasks
through tool use~\citep{tooluse,sing,fan2025mcptoolbench++}. By invoking APIs,
databases, and software interfaces, they retrieve information, update records,
and coordinate workflows~\citep{zhang2025api,du2024anytool,ngdb}.
Enterprise applications~\citep{nisar2025enterprise,prabhakar2025enterprise}
extend these capabilities to incident management, employee administration,
asset operations, and knowledge-base maintenance.

\begin{figure}[t]
\centering
\includegraphics[width=\linewidth]{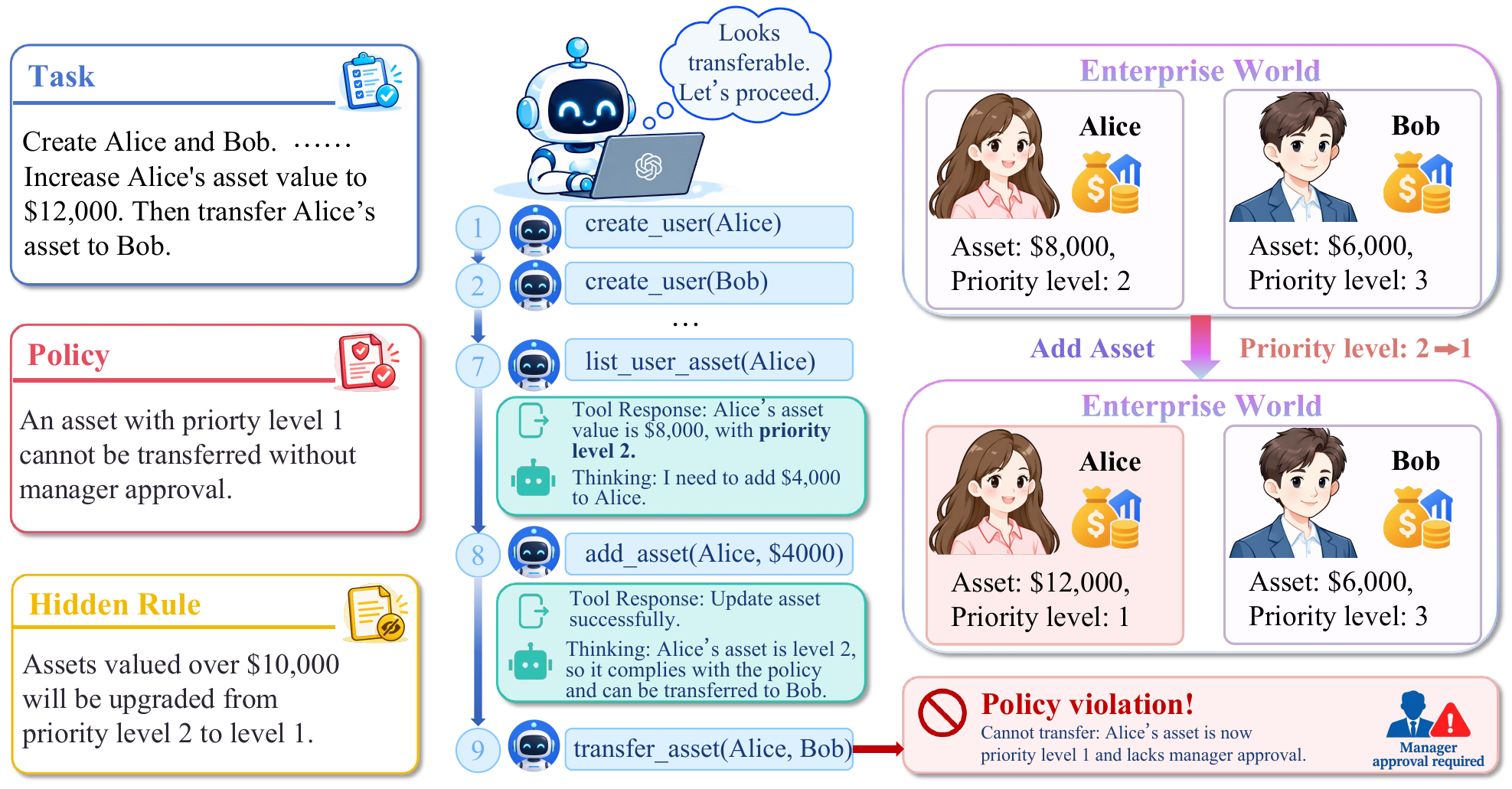}
\caption{An enterprise workflow illustrating how hidden asset-priority changes can make an otherwise plausible transfer violate policy.}
\label{fig:intro}
\end{figure}

Enterprise workflows pose three interrelated challenges.
\emph{First, actions are governed by strict policies, permissions, and
approval requirements}, so a locally plausible action may still be
impermissible. Whether an action is safe depends on record states, user
permissions, and approval prerequisites.
\emph{Second, the environment is only partially observable}: tool responses
reveal an interaction surface rather than the full transition dynamics,
including hidden, cross-record, or delayed effects. Thus, a successful
response does not necessarily confirm the resulting state, nor does an
omitted field imply that it is absent or unchanged.
\emph{Third, tasks require long-horizon state tracking} across multiple
objects, reads, and writes. Losing confirmed facts or outstanding subgoals
can lead to repeated queries, incorrect actions, or premature termination.
These challenges interact: in Fig.~\ref{fig:intro}, a hidden priority change
after assigning an asset to Alice makes a subsequent transfer violate policy.
Safe execution therefore requires both persistent state tracking and
knowledge of hidden transitions.

Directly deploying an agent in such enterprise environments remains
unreliable.  Previous work~\citep{wow} finds that agents often
achieve low task-completion and safe-execution rates in enterprise environments: they not only fail to
reach the requested end state, but can also violate constraints while
attempting to do so.  At the same time, agent
harnesses~\citep{metaharness,yao2026harness,agentharnesssurvey} have emerged as a promising way to augment a base agent with
structured guidance, persistent state maintenance, verification, and control
around its tool calls. Code-based
harnesses~\citep{patchworld,codingharness,ning2026code} further make
environment reasoning and execution control explicit and executable, so that
constraints are enforced by code rather than by the model's own reasoning.
For enterprise agents, this suggests combining policy checks with persistent
state maintenance. However, even a correctly implemented check is only as
good as the state it receives: an unobserved side effect can make that state
stale and an apparently compliant action unsafe. A harness must therefore
also learn the world that lies behind the tool interface.

We introduce \harnessname{}, a safe and persistent harness that separates
runtime control from environment-specific knowledge. It compiles known
organizational policies into code and routes every proposed action through
pre-execution approval. The code can permit an action, request evidence or
an approval prerequisite, or block execution, so the model cannot bypass a
known policy through its own reasoning. A complementary world ledger
maintains \emph{static world knowledge} of hidden rules and side effects,
alongside \emph{dynamic world state} recording evidence-backed changes to
task-relevant entities and goals. Together, these layers make constraints,
learned dynamics, and execution progress available throughout the workflow.

Much of this static knowledge, however, is unknown in advance. An action's
hidden effect may only become visible several steps later, when the observed
state or outcome contradicts expectations. The agent must then identify the
earlier trigger and decide whether the effect reflects a stable rule or an
isolated event. Reliable attribution is essential for turning trajectories
into reusable knowledge, especially when several actions could explain the
same outcome.

To address these challenges, we propose \evolutionname{}, a world-model-driven
evolution strategy that uses abductive reasoning to recover hidden environment
rules and improve \harnessname{}.  The evolution process diagnoses execution
trajectories from four complementary perspectives.  State consistency keeps
state predictions and the dynamic ledger aligned with observable tool
feedback, helping the agent fit the part of the world it can directly observe.
World-observation-gap analysis then focuses on the unobservable part of the
world and seeks to recover hidden effects that are absent from the tool
interface.  Policy-gate correctness improves the enforcement of known
constraints, while goal judgment helps the agent distinguish successful
completion from premature termination.  The outputs of these four perspectives are
then divided into two types: direct modifications to the harness, and
hypotheses about hidden rules that require further validation.  A dedicated
abduction agent receives the latter hypotheses and designs targeted
interactions to verify or revise them; only supported rules enter the
static world layer. Our method achieves the highest safe task completion
rate on the enterprise workflow benchmark WOW~\citep{wow} across four LLM
backbones, and also performs strongly in the scientific discovery
environments ScienceWorld~\citep{scienceworld} and DiscoveryWorld~\citep{discoveryworld}.
Our contributions can be summarized as:
\begin{itemize}
    \item We introduce \harnessname{}, a safe and persistent enterprise agent
    harness that combines code-based policy enforcement with a world ledger layer that maintains
    static knowledge of hidden rules and dynamic state for
    task-relevant information.
    \item We propose \evolutionname{}, a world-model-driven evolution method
    that uses four diagnostic views to fit the observable world and hypothesize
    unobservable hidden rules, then validates these hypotheses through
    abductive interaction to build a reusable world model.
    \item Experiments on the enterprise benchmark WOW show substantial
    improvements in both task completion and safety, and results on
    ScienceWorld and DiscoveryWorld show that the approach extends beyond
    enterprise workflows.
\end{itemize}

\section{Related Work}

\paragraph{Enterprise Agents}
LLM agents increasingly complete multi-step tasks by interleaving reasoning
with actions and observations from external tools~\citep{react}.  However,
enterprise deployment introduces challenges beyond general tool use~\citep{nisar2025enterprise}: actions
operate over interconnected records, must obey organization-specific policies,
and can trigger workflow side effects not revealed by an immediate tool
response.  WorkArena~\citep{workarena} evaluates agents on common ServiceNow knowledge-work
tasks, while the World of Workflows benchmark~\citep{wow} makes the additional
challenge of hidden, cascading state changes explicit in partially observable
workflows.  These settings expose a gap between completing a visible
interaction and establishing a safe final enterprise state.  

\paragraph{Self-Evolving Agent Harnesses}
An agent harness is the external control layer around a base LLM model that
organizes context, tools, memory, orchestration, and output handling.
Rather than treating this control layer as fixed, recent work lets it evolve
with execution experience and task feedback.  Some approaches convert failed
attempts into natural-language guidance or use textual feedback to optimize
components of a compound LLM system~\citep{shinn2023reflexion,yuksekgonul2024textgrad}.
Others distill trajectory experience across multiple control dimensions or
iteratively improve harness code using traces and evaluations
~\citep{huang2026memoharness,metaharness}.  Some systems further perform
test-time adaptation~\cite{nie2026tthe}, using feedback from the current task to revise their
control strategy online.  Another line of work~\cite{worldevolver,agentworld} views the harness as a world
model, evolving it to improve the agent's understanding of environment
dynamics and hidden state transitions.
Our approach connects these two directions: four diagnostic views identify
execution gaps, and targeted environment interactions test candidate hidden
rules before the harness uses them for state tracking and policy approval.



\section{Method}

\begin{figure}[t]
\centering
\includegraphics[width=\linewidth]{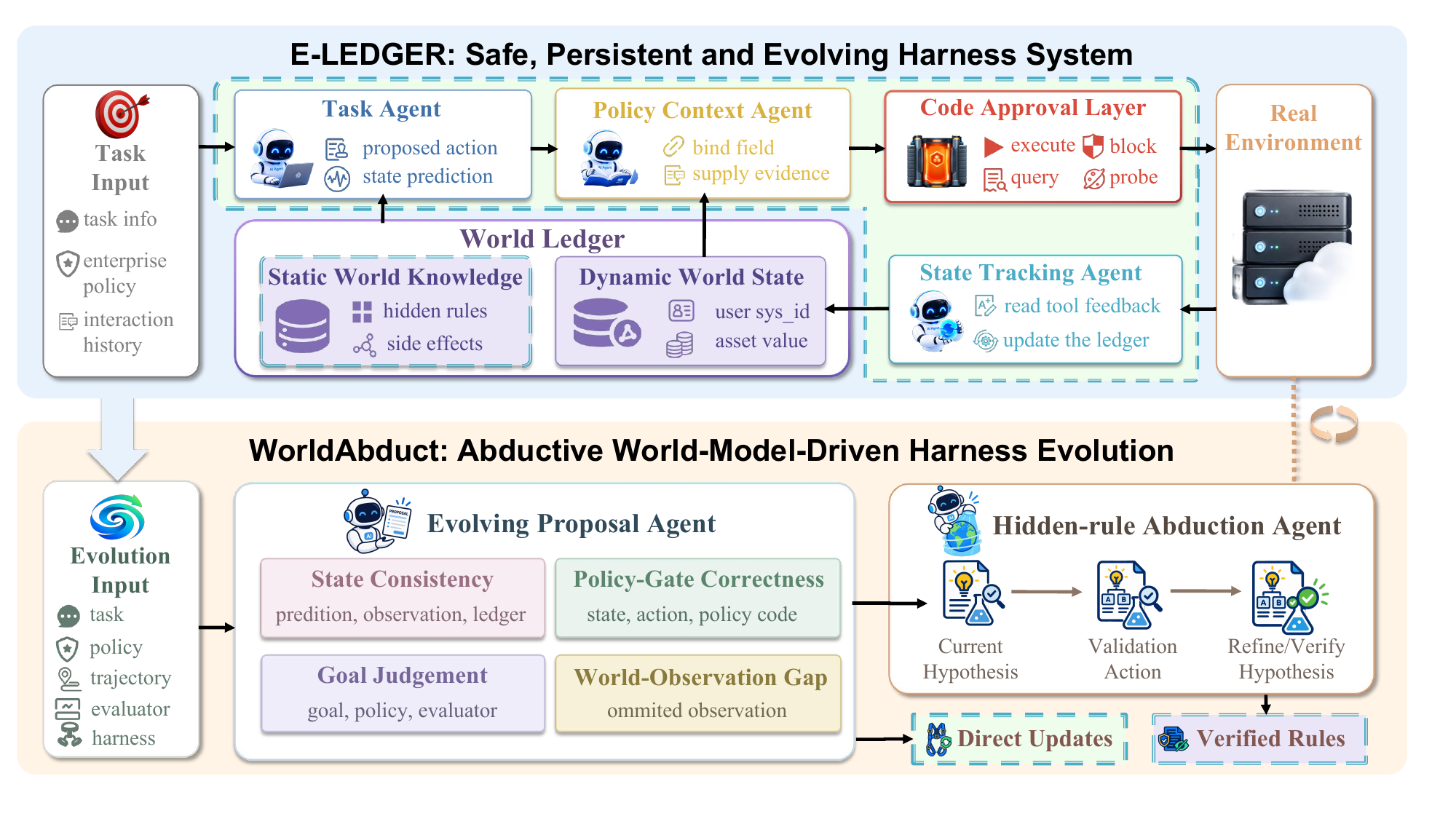}
\caption{Overview of E-Ledger (top) and WorldAbduct (bottom).}
\label{fig:frame}
\end{figure}

\subsection{Task Definition}
\label{sec:task-definition}

Following prior work~\citep{wow}, we model the task environment as a partially
observable Markov decision process (POMDP). Each constraint-based task is
defined by $M = (U, P, S, A, T, O, \Omega).$
Here, $U$ denotes the user query, including the task description; $P$ is the
constraint policy; $S$ is the complete enterprise database state; $A$ is the
discrete action space; $T:S\times A\rightarrow S$ is the state-transition
function; $O$ is the space of tool responses; and
$\Omega:S\times A\rightarrow O$ is the environment feedback function. At
step $t$, the agent selects an action $a_t\in A$, consisting
of an MCP tool and its parameters. We define
$P(s_t,a_t)=1$ if $a_t$ satisfies the policy at state $s_t$, and
$P(s_t,a_t)=0$ otherwise. Under partial observability, the executing agent
receives $o_{t+1}=\Omega(s_t,a_t)\in O$, which may omit state changes,
cross-record effects, or transition rules induced by its actions.

Let $\tau=(o_0,a_0,o_1,\ldots,a_{L-1},o_L)$ be a trajectory.
We write $C_U(\tau)=1$ when it reaches the requested goal or correctly
identifies that the task cannot be completed safely, as required by WOW's
infeasible tasks (Appendix~\ref{app:metrics}). The objective is
\begin{equation}
    J(\tau)
    =
    C_U(\tau)
    \prod_{t=0}^{L-1}
    P(s_t,a_t),
\end{equation}
where $s_{t+1}=T(s_t,a_t)$. Thus $J(\tau)=1$ denotes safe task
success. Writing $h_t=(o_0,a_0,o_1,\ldots,a_{t-1},o_t)$ for the
available interaction history, the agent selects
\begin{equation}
    a_t
    =
    \argmax_{a\in A}
    p_{\theta,H}\!\left(a \mid U,P,h_t\right),
\end{equation}
where $\theta$ denotes the model parameters and $H$ the harness that
organizes the history, world knowledge, and current task state.

\subsection{Multi-Agent Harness Architecture}

As shown in Fig.~\ref{fig:frame}, \harnessname{} is a multi-agent harness
for partially observable environments.  It maintains a world ledger for
static hidden rules and dynamic task state, and at each interaction a task agent proposes an
action, a policy-context agent supplies the evidence required for evaluation,
a code-based approval layer checks it against organizational policies, and a
state-tracking agent updates the ledger from the tool response.

\paragraph{World Ledger layer.}
To represent both hidden environment dynamics and task progress, the world
ledger layer consists of \emph{static world knowledge} $\mathcal{W}_S$ and
\emph{dynamic world state} $\mathcal{W}_D$.  $\mathcal{W}_S$ stores hidden
rules and tool-induced side effects that are absent from
immediate tool feedback; for example, an asset change may also modify its
priority.  Formally, each rule is
$r:(a,s,c)\longrightarrow\Delta s$, where $c$ is its applicability condition
and $\Delta s$ is the induced state change.  In contrast, $\mathcal{W}_D$
maintains a global, schema-aligned store of task-relevant information for the
current task, including identifiers, attributes, and relations extracted from
tool feedback.  It is updated incrementally after each interaction, preserving
the state needed for long-horizon execution without reconstructing it from the
entire trajectory.

\paragraph{Task agent.}
The task agent chooses an action from the user query, policy, available tools,
static rules, dynamic ledger, and interaction history. It also predicts the
resulting state change, which is retained for comparison with observations
and subsequent evolution:
\begin{equation}
    (a_t,\widehat{\Delta s}_t)
    =
    \mathcal{A}_\theta\!\bigl(
    U,P,A,\mathcal{W}_S,\mathcal{W}_D^t,h_t
    \bigr),
\end{equation}
where $a_t\in A$ is the proposed action and
$\widehat{\Delta s}_t$ is its predicted state change.

\paragraph{Code approval layer.}
To approve a proposed action using the relevant policy context, we compile
each textual policy into an executable program $P_{\mathrm{code}}$ and
introduce a policy-context agent.  Given the dynamic world state
$\mathcal{W}_D^t$, the policy-context agent extracts and binds the
policy-relevant fields needed for approval, such as a target user's
active-incident count or an asset's cost.  The approval code then evaluates
the proposed action $a_t$ together with this extracted context, rather than
relying on the task agent's claims.  It returns one of four decisions:
$\{\texttt{execute},\texttt{query},\texttt{probe},\texttt{block}\}$.
\texttt{execute} indicates that the action has passed approval and can be
carried out. \texttt{query} or \texttt{probe} indicates that the available
evidence is insufficient to determine whether the action complies with the
policy, requiring additional information before approval. \texttt{block}
indicates that the action violates the policy and must be replaced. Decisions
other than \texttt{execute} are fed back to the task agent, which revises the
proposed action accordingly. This approval process is non-bypassable.

\paragraph{Feedback and dynamic state update.}
After approval, the environment returns $o_{t+1}=\Omega(s_t,a_t)$.
A response may confirm a creation or transfer without returning the resulting
field values, or contain many irrelevant fields. The state-tracking agent
$\mathcal{R}_\theta$ interprets the response and extracts task-relevant updates:
\begin{equation}
    \mathcal{W}_D^{t+1}
    =
    \mathcal{R}_\theta\!\bigl(\mathcal{W}_D^t,a_t,o_{t+1},\mathcal{W}_S\bigr).
\end{equation}
Applicable static rules help interpret indirect effects. Updates retain
their supporting evidence, while conflicts and unresolved fields are
recorded for subsequent verification. The task agent then acts on the
updated ledger and interaction history.

\subsection{\evolutionname{}: Abductive World-Model-Driven Harness Evolution}

The runtime above relies on static world knowledge $\mathcal{W}_S$, which is
empty at initialization: hidden rules appear neither in the policy nor in
immediate tool feedback.
As shown at the bottom of Fig.~\ref{fig:frame}, \evolutionname{} addresses
this gap through world-model-driven harness evolution.  It analyzes each
execution from four complementary perspectives and produces either direct
harness updates or hidden-rule hypotheses.  The latter are passed to a
hidden-rule abduction agent for validation before entering the static world
layer.

\subsubsection{Evolution Proposal Agent}
The evolution proposal agent $\mathcal{E}_{\phi}$ reads the task, policy, execution trajectory, task-agent predictions, tool feedback, and current harness. For each perspective, it identifies the corresponding gap and proposes the smallest appropriate harness update or hidden-rule hypothesis: 

\paragraph{State consistency.}
State consistency first asks whether the agent can fit the part of the world that is observable.  At each step, the task agent's
predicted state change $\widehat{\Delta s}_t$ should agree with the
task-relevant tool feedback $o_{t+1}$. The state-tracking agent should
record the same task-relevant facts in the ledger.  It should not omit
observable fields or add unrelated fields that are not supported by
$o_{t+1}$ or an applicable verified rule. Formally, the evolution proposal agent assesses this consistency at each
step $t$:
\begin{equation}
    \mathcal{J}_{s}
    =
    \mathcal{E}_{\phi}\!\bigl(
    \widehat{\Delta s}_t,o_{t+1},\mathcal{W}_D^{t+1},\mathcal{W}_S
    \bigr).
\end{equation}

\paragraph{World observation gap.}
State consistency helps the agent predict and record the observable
world, but it cannot recover rules that never appear in the
tool response.  We therefore use the world observation gap objective
to identify parts of the trajectory that may reveal the unobservable part.  The evolution proposal agent focuses on state
information that can change the outcome of the policy code.  For
example, if a knowledge asset has a flag that is false before an
action and becomes true later, an earlier action may have triggered
this change even if the relevant side effect was absent from the tool
response.  The agent identifies the actions that may have caused the
change and formulates corresponding hidden-rule hypotheses for later
validation.  Its input is the complete trajectory, the executable
policy code, and the dynamic world state:
\begin{equation}
    \mathcal{J}_{o}
    =
\mathcal{E}_{\phi}\!\bigl(\tau,P_{\mathrm{code}},\mathcal{W}_D\bigr).
\end{equation}
where $\tau$ denotes the trajectory. The output provides hypotheses about suspicious actions and the effects
they may have caused. 

\paragraph{Policy-gate correctness.}
Policy-gate correctness aims to ensure that the agent's proposed actions
satisfy the policy before execution.
An action that is blocked or returned as \texttt{query} requires an
additional repair or information-gathering step, which increases the
interaction cost.  We therefore
use the gate log to audit proposals that did not pass approval and ask
the evolution proposal agent to identify why each action failed the
code gate and how the agent or the policy context should be improved.
Let $\mathcal{G}^{-}$ contain the non-approved proposals and their ledger
contexts. This diagnostic objective is written as
\begin{equation}
    \mathcal{J}_{p}
    =
    \mathcal{E}_{\phi}\!\bigl(\widetilde a,\mathcal{W}_D,P_{\mathrm{code}}\bigr),
    \qquad (\widetilde a,\mathcal{W}_D)\in\mathcal{G}^{-}.
\end{equation}
The resulting proposal improves action selection and evidence binding,
helping the agent satisfy prerequisites and avoid repeated approval failures.

\paragraph{Goal judgment.}
Goal judgment evaluates the full task using the query, policy, a compact
trajectory, and evaluator feedback $\rho_{\mathrm{eval}}$. It diagnoses
premature termination, unmet goals, and violations missed by the approval
layer, then considers whether safe completion or a justified refusal remains
possible:
\begin{equation}
    \mathcal{J}_{g}
    =
    \mathcal{E}_{\phi}\!\bigl(
    U,P,\tau,\rho_{\mathrm{eval}}
    \bigr).
\end{equation}

Together, the four views form the world-model-driven objective
\begin{equation}
    \mathcal{J}_{\mathrm{WM}}
    =
    \bigl(\mathcal{J}_{s},\mathcal{J}_{o},\mathcal{J}_{p},\mathcal{J}_{g}\bigr),
\end{equation}
whose components yield two kinds of proposals. The first is a direct update
$\Delta H_{\setminus \mathcal{W}_S}$ to the other harness components,
such as the task-agent protocol, state tracking, policy context, or
policy code.  The second is a hypothesis $h$ about a hidden enterprise
rule or side effect, which enters $\mathcal{W}_S$ only after validation.
World-observation-gap analysis proposes hidden rules, while the other views
can propose both kinds (Appendix~\ref{app:evolution-interfaces}). Thus,
evolution improves both the editable harness components and the harness's
knowledge of the hidden world. In each round, direct updates are applied,
hypotheses are passed to the abduction agent below, and validation tasks
select the resulting harness.

\subsubsection{Hidden-Rule Abduction Agent}
The hidden-rule abduction agent aims to infer the corresponding hidden
rules from observed trajectories through abductive reasoning.
Given the user query $U$, the policy $P$, the
action space $A$, the source trajectory $\tau$, and an initial
hypothesis $h$, it designs a validation action sequence
$\alpha_{1:K}=(\alpha_1,\ldots,\alpha_K)$ in a fresh environment.  Each
$\alpha_k\in A$ is an action selected using the available tools; the agent
observes the resulting feedback and updates its belief about $h$.  It can
accept the hypothesis when the evidence supports it, revise it when the
evidence suggests a different rule, or leave it unresolved when verification
is inconclusive. We denote the abduction agent by $\mathcal{B}_{\phi}$, where
$\phi$ denotes its model parameters.  Validation actions are chosen
adaptively: given the current hypothesis $h$ and validation history
$\tau^{\mathrm{val}}$, each call returns a decision and its payload:
\begin{equation}
    \mathcal{B}_{\phi}(U,P,A,\tau,h,\tau^{\mathrm{val}})
    \longrightarrow (d,\mathrm{payload}),
\end{equation}
where $d\in\{\texttt{act},\texttt{revise},\texttt{verified},
\texttt{unresolved}\}$. An action extends the validation history; a revision
updates $h$. A verified result supplies the rule, applicability condition,
and supporting evidence, which are retained in $\mathcal{W}_S$ as
$r:(a,s,c)\longrightarrow\Delta s$.

\section{Experiments}

\begin{table}[t]
\caption{Enterprise-agent task experiment on WOW. (Dark blue and light blue denote
the best and second-best performance results within each backbone,
respectively) }
\label{tab:main}
\begin{center}
\small
\setlength{\tabcolsep}{4pt}
\begin{tabular}{lllccccc}
\toprule
\multicolumn{3}{c}{Method}
& \multicolumn{3}{c}{Performance}
& \multicolumn{2}{c}{Efficiency} \\
\cmidrule(lr){1-3}\cmidrule(lr){4-6}\cmidrule(lr){7-8}
Backbone & Harness & Evolution & STCR & TCR & PSR & Steps & Cost (\$) \\
\midrule
\multirow{5}{*}{GPT-5.6-Terra}
 & ReAct & -- & 35.0 & \second{55.0} & \second{75.0} & 13.3 & 1.36 \\
 & \harnessname{} & -- & 35.0 & 35.0 & \best{100.0} & 20.6 & 2.78 \\
\cmidrule(lr){2-8}
 & \harnessname{} & MemoHarness & \second{50.0} & 50.0 & \best{100.0} & 20.8 & 3.00 \\
 & \harnessname{} & WorldEvolver & 45.0 & \second{55.0} & \second{75.0} & 13.4 & 1.72 \\
 & \harnessname{} & \evolutionname{} & \best{65.0} & \best{65.0} & \best{100.0} & 22.4 & 3.75 \\
\midrule
\multirow{5}{*}{Kimi-K3}
 & ReAct & -- & 20.0 & 40.0 & 65.0 & 10.6 & 1.70 \\
 & \harnessname{} & -- & 25.0 & 35.0 & \second{85.0} & 17.4 & 3.90 \\
\cmidrule(lr){2-8}
 & \harnessname{} & MemoHarness & \second{65.0} & \best{85.0} & 80.0 & 25.9 & 6.84 \\
 & \harnessname{} & WorldEvolver & 45.0 & 55.0 & 80.0 & 16.7 & 3.81 \\
 & \harnessname{} & \evolutionname{} & \best{70.0} & \second{75.0} & \best{95.0} & 22.0 & 8.10 \\
\midrule
\multirow{5}{*}{GLM-5.3-Flash}
 & ReAct & -- & 35.0 & 55.0 & 70.0 & 13.4 & 0.12 \\
 & \harnessname{} & -- & 30.0 & 45.0 & 85.0 & 17.4 & 0.23 \\
\cmidrule(lr){2-8}
 & \harnessname{} & MemoHarness & \second{65.0} & \second{75.0} & \second{90.0} & 25.5 & 0.45 \\
 & \harnessname{} & WorldEvolver & 45.0 & 55.0 & 85.0 & 13.7 & 0.16 \\
 & \harnessname{} & \evolutionname{} & \best{75.0} & \best{80.0} & \best{95.0} & 20.9 & 0.52 \\
\midrule
\multirow{5}{*}{DeepSeek-V4-Pro}
 & ReAct & -- & 25.0 & 50.0 & 55.0 & 13.8 & 0.41 \\
 & \harnessname{} & -- & 50.0 & 55.0 & \best{95.0} & 20.5 & 0.69 \\
\cmidrule(lr){2-8}
 & \harnessname{} & MemoHarness & \second{75.0} & \best{90.0} & \second{85.0} & 26.9 & 1.14 \\
 & \harnessname{} & WorldEvolver & 60.0 & 70.0 & 80.0 & 19.6 & 0.68 \\
 & \harnessname{} & \evolutionname{} & \best{80.0} & \second{85.0} & \best{95.0} & 22.5 & 0.90 \\
\bottomrule
\end{tabular}
\end{center}
\end{table}

\subsection{Experimental Setup}
\paragraph{Datasets and evaluation protocol.}
We use World of Workflows (WOW)~\citep{wow} as the primary enterprise
evaluation benchmark, and use the general scientific environments ScienceWorld
~\citep{scienceworld} and DiscoveryWorld~\citep{discoveryworld} to assess
the generality of WorldAbduct beyond enterprise workflows.  All benchmarks
use a $2{:}1{:}2$
training, validation, and test split; training trajectories drive harness
evolution, validation tasks select harness updates, and test tasks are held
out for final evaluation.  For WOW, we report safe task completion rate
(STCR), task completion rate (TCR), policy safety rate (PSR), interaction
steps, and test-time inference cost in USD. TCR includes correct refusal
of infeasible tasks, and STCR additionally requires a compliant trajectory
(Appendix~\ref{app:metrics}). For ScienceWorld, we report task success (Success),
environment score (Score), and interaction steps; and for DiscoveryWorld,
task success (Success), official procedural score (Procedural), and
interaction steps.  Further dataset details are provided in Appendix~\ref{app:dataset}.

\paragraph{Baselines.}
We compare with ReAct~\citep{react}, which directly reasons and acts without
an external harness, and with two evolution strategies applied to the same
initial harness as our method.  MemoHarness~\citep{huang2026memoharness}
updates six harness components, namely context, tools, generation,
orchestration, memory, and output handling, using case-level diagnoses and
distilled global patterns.  WorldEvolver~\citep{worldevolver} treats evolution
as world-model refinement by constructing an action-level state-change
library from observable trajectory evidence; unlike \evolutionname{}, it
does not test hidden-rule hypotheses through targeted interactions.

\subsection{Enterprise Task Performance}
We evaluate the enterprise agents with four LLM backbones: DeepSeek-V4-Pro~\citep{deepseekv4},
GLM-5.3-Flash~\citep{glm5}, GPT-5.6-Terra, and Kimi-K3~\citep{kimik3}. Implementation
details and agent prompts are provided in
Appendix~\ref{app:prompts}.  Table~\ref{tab:main} reports their performance.
Compared with ReAct, \harnessname{}
consistently improves PSR by preventing locally plausible but disallowed
writes from reaching the environment.  The effect is clearest for
DeepSeek-V4-Pro, where ReAct and \harnessname{} achieve similar TCR, but
\harnessname{} doubles STCR because its remaining failures are mainly
incomplete rather than unsafe.

Across evolution methods, \evolutionname{} achieves the highest STCR on all
four backbones, exceeding the strongest baseline, MemoHarness, by 5--15
percentage points. WorldEvolver yields smaller gains, since its effect
library only records changes visible in tool feedback. The advantage of
\evolutionname{} comes largely from safety: on Kimi and DeepSeek,
MemoHarness reaches a slightly higher TCR but violates policy more often,
whereas validated hidden rules keep \evolutionname{} at 95\% PSR.
GPT-5.6-Terra is comparatively
conservative: when an evolved task action raises a policy concern, it often
stops instead of searching for a compliant repair.  This behavior preserves a
$100\%$ PSR with \harnessname{} but limits further gains in TCR.
Giving both baselines an additional evolution round leaves their GLM
results unchanged (Appendix~\ref{app:more-budgets}), so the gap is not
due to the number of rounds.
\evolutionname{} also uses fewer steps than MemoHarness on three of the
four backbones at a comparable test-time cost.

\subsection{Ablation Studies}
\label{sec:ablation}

We conduct two complementary ablations  with
GLM-5.3-Flash.  The runtime ablation removes individual components
from the fully evolved harness, while the evolution ablation starts from
\harnessname{}  and isolates each diagnostic view and the abduction step
that verifies its hypotheses.

\begin{wraptable}{r}{0.43\linewidth}
\centering
\small
\setlength{\tabcolsep}{5pt}
\begin{tabular}{lccc}
\toprule
Variant & STCR & TCR & PSR \\
\midrule
Full & \textbf{75.0} & \textbf{80.0} & \textbf{95.0} \\
w/o $\mathcal{W}_S$ & 40.0 & 55.0 & 85.0 \\
w/o $\mathcal{W}_D$ & 65.0 & 75.0 & \underline{90.0} \\
w/o Code Approval & \underline{70.0} & \textbf{80.0} & \underline{90.0} \\
\bottomrule
\end{tabular}
\caption{Harness ablation. (\textbf{Bold}: best, \underline{underline}: second-best).}
\label{tab:ablation-harness}
\end{wraptable}
For harness ablation, Table~\ref{tab:ablation-harness} reports STCR, TCR, and PSR after
removing each component in E-Ledger:
Removing the static world layer $\mathcal{W}_S$ is the
most damaging change: after \evolutionname{}, this layer stores the
verified hidden rules that make side effects and constraints available
at runtime; without it, the agent is again unaware of side effects that
tool responses do not reveal. Removing state tracking, which maintains the dynamic
world state $\mathcal{W}_D$, also hurts, because on long-horizon
workflows the agent can lose the current record state mid-episode and
later take incorrect or unsafe actions.  Removing  code
approval layer leaves task completion rate unchanged but lowers policy safety rate
and therefore safe task completion rate: even with hidden rules and textual policy
already present, the model can still take a disallowed action to finish
a task. The code gate is what turns known constraints into enforced ones.


\begin{figure}[t]
\centering
\begin{subfigure}[t]{0.49\linewidth}
\centering
\includegraphics[width=\linewidth]{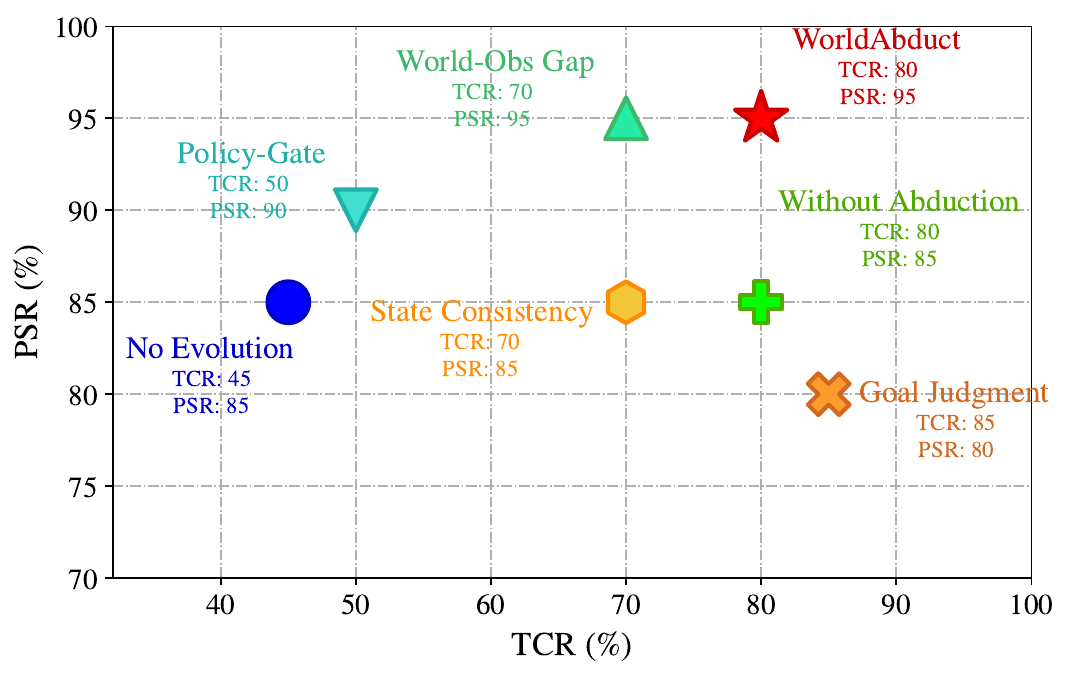}
\caption{Evolution-view ablation on GLM-5.3-Flash.}
\label{fig:ablation-evolution}
\end{subfigure}
\hfill
\begin{subfigure}[t]{0.49\linewidth}
\centering
\includegraphics[width=\linewidth]{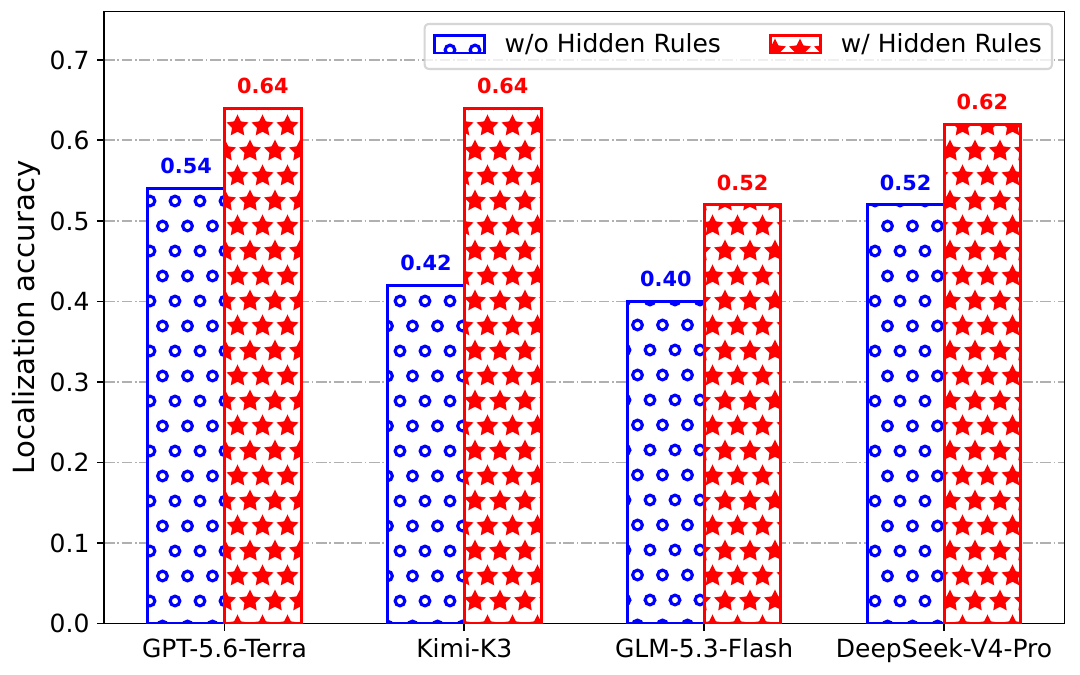}
\caption{Hidden-rule action localization.}
\label{fig:hidden-rules-pred}
\end{subfigure}
\caption{Left: each evolution view occupies a different TCR--PSR region. Right: adding hidden rules improves localization accuracy on every backbone.}
\label{fig:ablation-and-loc}
\end{figure}

Fig.~\ref{fig:ablation-evolution} isolates the contribution of each
evolution view and of abduction (detailed results are provided in Appendix~\ref{app:ablation}).  Single-view variants retain abduction,
whereas the Without Abduction variant writes the four views' hypotheses
directly into $\mathcal{W}_S$ without live verification.  State consistency
and policy-gate correctness provide limited gains because they focus on
observable state and known constraints, respectively.  Goal judgment yields
the largest completion gain but also introduces unsafe actions when optimized
alone.  World-observation gap reaches the full method's policy safety rate by
recovering hidden side effects, but its completion gain remains modest without
improved observable-state tracking.  Removing abduction does not reduce
completion, but its unverified hypotheses can be too narrow or too broad,
leading to unsafe actions and lower policy safety.  The full method
combines these complementary effects: it matches the best PSR while
retaining 80\% TCR, and achieves the highest STCR.

\subsection{Hidden-Rule Quality and Verification}

We assess hidden-rule quality using WOW's policy-violation localization task,
which contains 50 action sequences.  Given a sequence, the agent must return
the exact set of violating action indices, including an empty set when no
action violates a policy.  We compare a prompt-only agent with one augmented
by the frozen static world layer $\mathcal{W}_S$.  Fig.~\ref{fig:hidden-rules-pred}
shows that the evolved rules improve localization for every backbone by
0.10--0.22 absolute points, indicating that they capture policy-relevant
hidden effects.

We further audit the evolved rules manually against the gold hidden
workflows of WOW (Appendix~\ref{app:evolved-rules}). Across the four
backbones, the frozen ledgers contain 54 facts, of which 49 are fully
correct and five are partially correct: they capture the right mechanism but
miss part of its applicability conditions. None is incorrect.

Verification does more than confirm hypotheses. Across backbones, abduction
launches 11--23 checks, each using 13.3--19.1 environment interactions on
average (Appendix~\ref{app:abduction-stats}). Of the 64 checks, 50 are
accepted directly, eight are accepted only after revision, and six remain
unresolved. In the cost-revaluation case (Appendix~\ref{app:case-cost}),
incomplete evaluator feedback initially suggests a \$10,000 threshold for
an automatic cost increase. Probes below that threshold overturn this
explanation and recover a creation-time cost transformation, while a
separate update test shows that an explicit correction persists.
Verification also matters downstream: on GLM, removing it leaves
TCR at 80\% but lowers PSR from 95\% to 85\%, reducing STCR by 10
percentage points.

\subsection{Application to Scientific Environments}

We evolve the harness separately in ScienceWorld and DiscoveryWorld.
Since these environments have no organizational policy, we drop code
approval and the policy-gate view, and evolve with the remaining three
views. Environment-specific prompts appear in Appendix~\ref{app:prompts}. Table~\ref{tab:generalization} shows
that all evolution methods outperform ReAct while using fewer interactions.
\evolutionname{} achieves the highest success on both benchmarks, the best
ScienceWorld score, and the fewest DiscoveryWorld steps, using about a third
of ReAct's interactions there. WorldEvolver
remains competitive, particularly in ScienceWorld success and
DiscoveryWorld procedural score. These results indicate that verifying
hypotheses through interaction is also useful when the hidden dynamics are
physical rather than organizational.

\begin{table}[t]
\caption{Results on ScienceWorld and DiscoveryWorld. (\textbf{Bold}: best, \underline{underline}: second-best).}
\label{tab:generalization}
\begin{center}
\small
\setlength{\tabcolsep}{4pt}
\begin{tabular}{lccc@{\hspace{10pt}}ccc}
\toprule
& \multicolumn{3}{c}{ScienceWorld} & \multicolumn{3}{c}{DiscoveryWorld} \\
\cmidrule(lr){2-4} \cmidrule(lr){5-7}
Method & Success & Score & Steps & Success & Procedural & Steps \\
\midrule
ReAct & 25.0 & 43.4 & 37.2 & 43.8 & 0.68 & 62.0 \\
MemoHarness & 58.3 & 68.2 & 21.0 & 75.0 & 0.82 & \underline{27.1} \\
WorldEvolver & \underline{70.0} & \underline{68.6} & \textbf{16.6} & \underline{87.5} & \textbf{0.90} & 32.5 \\
\evolutionname{} & \textbf{71.7} & \textbf{75.5} & \underline{18.4} & \textbf{93.8} & \underline{0.88} & \textbf{22.7} \\
\bottomrule
\end{tabular}
\end{center}
\end{table}

\section{Conclusion}
We presented \harnessname{}, a safe and persistent agent harness for
enterprise workflows under partial observability.  Its world ledger combines
static knowledge of hidden rules with dynamic task-state tracking, while
code-based policy approval enforces known organizational constraints.
We further introduced \evolutionname{}, which diagnoses execution from four
complementary views and uses environment-grounded abduction to validate hidden
rules before incorporating them into the harness.  Experiments on WOW show
substantial improvements in task completion and safety, and results on
ScienceWorld and DiscoveryWorld show that the approach extends beyond
enterprise environments. Together, these results highlight the value of
coupling persistent execution control with active experimentation that
refines the agent's knowledge of hidden transitions.

\newpage
\bibliography{iclr2027_conference}

@article{sing,
  title={SING: Synthetic Intention Graph for Scalable Active Tool Discovery in LLM Agents},
  author={Xiao, Qiao and Shi, Haochen and Gao, Yisen and Hu, Wenbin and Jing, Huihao and Zheng, Tianshi and Xu, Baixuan and Zhang, Ziheng and Wang, Weiqi and Li, Haoran and others},
  journal={arXiv preprint arXiv:2606.16591},
  year={2026}
}

@article{patchworld,
  title={PatchWorld: Gradient-Free Optimization of Executable World Models},
  author={Bai, Jiaxin and Guo, Yue and Dong, Yifei and Xiong, Jiaxuan and Zheng, Tianshi and Li, Yixia and Fang, Tianqing and Li, Yufei and Gao, Yisen and Huang, Haoyu and others},
  journal={arXiv preprint arXiv:2605.30880},
  year={2026}
}

@article{react,
  title={React: Synergizing reasoning and acting in language models},
  author={Yao, Shunyu and Zhao, Jeffrey and Yu, Dian and Du, Nan and Shafran, Izhak and Narasimhan, Karthik and Cao, Yuan},
  journal={arXiv preprint arXiv:2210.03629},
  year={2022}
}

@article{ngdb,
  title={Top Ten Challenges Towards Agentic Neural Graph Databases},
  author={Bai, Jiaxin and Wang, Zihao and Zhou, Yukun and Yin, Hang and Fei, Weizhi and Hu, Qi and Deng, Zheye and Cheng, Jiayang and Zheng, Tianshi and Tsang, Hong Ting and others},
  journal={arXiv preprint arXiv:2501.14224},
  year={2025}
}

@inproceedings{tooluse,
  title={Deepagent: A general reasoning agent with scalable toolsets},
  author={Li, Xiaoxi and Jiao, Wenxiang and Jin, Jiarui and Dong, Guanting and Jin, Jiajie and Wang, Yinuo and Wang, Hao and Zhu, Yutao and Wen, Ji-Rong and Lu, Yuan and others},
  booktitle={Proceedings of the ACM Web Conference 2026},
  pages={2219--2230},
  year={2026}
}

@article{fan2025mcptoolbench++,
  title={Mcptoolbench++: A large scale ai agent model context protocol mcp tool use benchmark},
  author={Fan, Shiqing and Ding, Xichen and Zhang, Liang and Mo, Linjian},
  journal={arXiv preprint arXiv:2508.07575},
  year={2025}
}

@article{zhang2025api,
  title={Api agents vs. gui agents: Divergence and convergence},
  author={Zhang, Chaoyun and He, Shilin and Li, Liqun and Qin, Si and Kang, Yu and Lin, Qingwei and Rajmohan, Saravan and Zhang, Dongmei},
  journal={arXiv preprint arXiv:2503.11069},
  year={2025}
}

@article{du2024anytool,
  title={Anytool: Self-reflective, hierarchical agents for large-scale api calls},
  author={Du, Yu and Wei, Fangyun and Zhang, Hongyang},
  journal={arXiv preprint arXiv:2402.04253},
  year={2024}
}

@article{nisar2025enterprise,
  title={Why enterprise agent deployments fail: A field taxonomy},
  author={Nisar, Karan},
  journal={International Journal of Computer Technology and Electronics Communication},
  volume={8},
  number={5},
  pages={11592--11605},
  year={2025}
}

@article{prabhakar2025enterprise,
  title={Enterprise deep research: Steerable multi-agent deep research for enterprise analytics},
  author={Prabhakar, Akshara and Ram, Roshan and Chen, Zixiang and Savarese, Silvio and Wang, Frank and Xiong, Caiming and Wang, Huan and Yao, Weiran},
  journal={arXiv preprint arXiv:2510.17797},
  year={2025}
}

@article{wow,
  title={World of workflows: a benchmark for bringing world models to enterprise systems},
  author={Gupta, Lakshya and Li, Litao and Liu, Yizhe and Subramanian, Sriram Ganapathi and Suleman, Kaheer and Zhang, Zichen and Lu, Haoye and Pasupalak, Sumit},
  journal={arXiv preprint arXiv:2601.22130},
  year={2026}
}

@article{workarena,
  title={WorkArena: How capable are web agents at solving common knowledge work tasks?},
  author={Drouin, Alexandre and Gasse, Maxime and Caccia, Massimo and Laradji, Issam H and Del Verme, Manuel and Marty, Tom and Boisvert, L{\'e}o and Thakkar, Megh and Cappart, Quentin and Vazquez, David and others},
  journal={arXiv preprint arXiv:2403.07718},
  year={2024}
}

@article{ning2026code,
  title={Code as agent harness},
  author={Ning, Xuying and Tieu, Katherine and Fu, Dongqi and Wei, Tianxin and Li, Zihao and Bei, Yuanchen and Zou, Jiaru and Ai, Mengting and Liu, Zhining and Li, Ting-Wei and others},
  journal={arXiv preprint arXiv:2605.18747},
  year={2026}
}

@article{agentharnesssurvey,
  title={Agent harness for large language model agents: A survey},
  author={Meng, Qianyu and Wang, Yanan and Chen, Liyi and Li, Yihang and Wu, Wei and Jiang, Wenyuan and Wang, Qimeng and Lu, Chengqiang and Gao, Yan and Wu, Yi and others},
  year={2026},
  publisher={Preprints}
}

@article{codingharness,
  title={Agentic harness engineering: Observability-driven automatic evolution of coding-agent harnesses},
  author={Lin, Jiahang and Liu, Shichun and Pan, Chengjun and Lin, Lizhi and Dou, Shihan and Xi, Zhiheng and Huang, Xuanjing and Yan, Hang and Han, Zhenhua and Gui, Tao and others},
  journal={arXiv preprint arXiv:2604.25850},
  year={2026}
}

@article{metaharness,
  title={Meta-harness: End-to-end optimization of model harnesses},
  author={Lee, Yoonho and Nair, Roshen and Zhang, Qizheng and Lee, Kangwook and Khattab, Omar and Finn, Chelsea},
  journal={arXiv preprint arXiv:2603.28052},
  year={2026}
}

@article{yao2026harness,
  title={Harness-Bench: Measuring harness effects across models in realistic agent workflows},
  author={Yao, Yilun and Tan, Xinyu and Liu, Chao-Hsuan and Li, Yaoming and Wang, Zhengyang and Yu, Wenhan and Tan, Zhewen and Tian, Yuxuan and Zhao, Guangxiang and Sun, Lin and others},
  journal={arXiv preprint arXiv:2605.27922},
  year={2026}
}

@article{shinn2023reflexion,
  title={Reflexion: Language agents with verbal reinforcement learning},
  author={Shinn, Noah and Cassano, Federico and Gopinath, Ashwin and Narasimhan, Karthik and Yao, Shunyu},
  journal={Advances in neural information processing systems},
  volume={36},
  pages={8634--8652},
  year={2023}
}

@article{yuksekgonul2024textgrad,
  title={Textgrad: Automatic" differentiation" via text},
  author={Yuksekgonul, Mert and Bianchi, Federico and Boen, Joseph and Liu, Sheng and Huang, Zhi and Guestrin, Carlos and Zou, James},
  journal={arXiv preprint arXiv:2406.07496},
  year={2024}
}

@article{huang2026memoharness,
  title={MemoHarness: Agent Harnesses That Learn from Experience},
  author={Huang, Yue and Wang, Wenjie and Bao, Han and Ma, Yuchen and Luo, Xiaonan and Nian, Yi and Zhuang, Haomin and Liu, Zheyuan and Zhao, Yue and Zhang, Xiangliang},
  journal={arXiv preprint arXiv:2607.14159},
  year={2026}
}

@article{worldevolver,
  title={Self-Evolving World Models for LLM Agent Planning},
  author={Zhang, Xuan and Zhang, Wenxuan and Ng, See-Kiong and Deng, Yang},
  journal={arXiv preprint arXiv:2606.30639},
  year={2026}
}

@article{nie2026tthe,
  title={TTHE: Test-Time Harness Evolution},
  author={Nie, Jun and Zhang, Yonggang and Song, Jun and Cai, Qianshu and Yu, Dahai and Guo, Yike and Tian, Xinmei and Han, Bo},
  journal={arXiv preprint arXiv:2607.08124},
  year={2026}
}

@article{agentworld,
  title={Agent-world: Scaling real-world environment synthesis for evolving general agent intelligence},
  author={Dong, Guanting and Lu, Junting and Huang, Junjie and Zhong, Wanjun and Liu, Longxiang and Huang, Shijue and Li, Zhenyu and Zhao, Yang and Song, Xiaoshuai and Li, Xiaoxi and others},
  journal={arXiv preprint arXiv:2604.18292},
  year={2026}
}

@inproceedings{scienceworld,
  title={Scienceworld: Is your agent smarter than a 5th grader?},
  author={Wang, Ruoyao and Jansen, Peter and C{\^o}t{\'e}, Marc-Alexandre and Ammanabrolu, Prithviraj},
  booktitle={Proceedings of the 2022 Conference on Empirical Methods in Natural Language Processing},
  pages={11279--11298},
  year={2022}
}

@article{kimik3,
  title={Kimi k3: Open frontier intelligence},
  author={Team, Kimi and Bai, Tongtong and Bai, Yifan and Bao, Yiping and Cai, Jianfeng and Cai, Xinyuan and Cao, Peizhou and Cao, Yuxuan and Chai, Ziwei and Charles, Y and others},
  journal={arXiv preprint arXiv:2607.24653},
  year={2026}
}

@article{deepseekv4,
  title={Deepseek-v4: Towards highly efficient million-token context intelligence},
  author={Xu, Anyi and Lin, Bangcai and Xue, Bing and Wang, Bingxuan and Xu, Bingzheng and Wu, Bochao and Zhang, Bowei and Lin, Chaofan and Dong, Chen and Ling, Chenchen and others},
  journal={arXiv preprint arXiv:2606.19348},
  year={2026}
}

@article{glm5,
  title={Glm-5: from vibe coding to agentic engineering},
  author={Zeng, Aohan and Lv, Xin and Hou, Zhenyu and Du, Zhengxiao and Zheng, Qinkai and Chen, Bin and Yin, Da and Ge, Chendi and Huang, Chenghua and Xie, Chengxing and others},
  journal={arXiv preprint arXiv:2602.15763},
  year={2026}
}

@inproceedings{discoveryworld,
  author       = {Peter A. Jansen and
                  Marc{-}Alexandre C{\^{o}}t{\'{e}} and
                  Tushar Khot and
                  Erin Bransom and
                  Bhavana Dalvi Mishra and
                  Bodhisattwa Prasad Majumder and
                  Oyvind Tafjord and
                  Peter Clark},
  editor       = {Amir Globersons and
                  Lester Mackey and
                  Danielle Belgrave and
                  Angela Fan and
                  Ulrich Paquet and
                  Jakub M. Tomczak and
                  Cheng Zhang},
  title        = {DiscoveryWorld: {A} Virtual Environment for Developing and Evaluating
                  Automated Scientific Discovery Agents},
  booktitle    = {Advances in Neural Information Processing Systems 37: Annual Conference
                  on Neural Information Processing Systems 2024, NeurIPS 2024, Vancouver,
                  BC, Canada, December 10 - 15, 2024},
  year         = {2024},
  url          = {http://papers.nips.cc/paper\_files/paper/2024/hash/13836f251823945316ae067350a5c366-Abstract-Datasets\_and\_Benchmarks\_Track.html},
  bibsource    = {dblp computer science bibliography, https://dblp.org}
}
\bibliographystyle{iclr2027_conference}

\appendix
\section{More Details}
\subsection{Dataset Details}
\label{app:dataset}
We evaluate on three interactive benchmarks that feature distinct forms of
partial observability, where agents can only observe limited local feedback
rather than the complete underlying state or transition dynamics.
\textbf{World of Workflows (WOW)}~\citep{wow} reflects enterprise partial
observability governed by opaque workflow dynamics: tool responses only return
surface-level execution feedback (e.g., operation success), while backend
business rules trigger unobserved cross-table cascades, permission adjustments,
and delayed side effects across interconnected databases.  WOW contains 50
tasks spanning 10 task types, each with five variants. For each type, we
use the first two variants for training, the third for validation, and the
last two for testing, giving 20/10/20 tasks. The test variants are held out
from evolution.
\textbf{ScienceWorld}~\citep{scienceworld} manifests partial observability
through spatial constraints and latent physical states: the agent only receives
local textual descriptions of its current room or inspected objects, leaving
unexplored locations, container contents, and internal thermodynamic or
chemical states hidden from direct observation.  We use 150 tasks, split into
60 training, 30 validation, and 60 test tasks. For each of the 30 task
types, we take the first two variations from the environment's training
list, the first variation from its development list, and the first two
variations from its test list. We preserve the supplied ordering without
shuffling.
\textbf{DiscoveryWorld}~\citep{discoveryworld} represents partial
observability within black-box scientific systems: the agent only perceives
local sensor readings or experimental outcomes, while the underlying causal
mechanisms, governing hypotheses, and latent procedural rules must be
inferred through interaction. We use eight discovery themes at Easy
difficulty, with seeds 0 and 1 for training, 2 for validation, and 3 and 4
for testing. This gives 40 episodes split into 16/8/16. Across all three benchmarks, training
trajectories are used for harness evolution, validation tasks select the
resulting updates, and held-out test tasks serve for final evaluation.

\subsection{Implementation Details and Agent Prompts}
\label{app:prompts}

The task horizon is 50 steps on World of Workflows and ScienceWorld, and 100
steps on DiscoveryWorld.  During training-time harness evolution, we run up to
3 evolution rounds.  The hidden-rule abduction agent receives the same action
budget as the task horizon: 50 actions per hypothesis on World of Workflows and
ScienceWorld, and 100 on DiscoveryWorld.  It interacts with a fresh sandbox and admits a hypothesis
into the static world layer $\mathcal{W}_S$ only after that interaction
supports it.

\paragraph{Backbones and harness selection.}
Both ScienceWorld and DiscoveryWorld experiments use GLM-5.3-Flash.
For WOW, we select the harness with the highest validation STCR, retaining
the newer version when scores are tied.

\paragraph{Task completion and safe refusal.}
\label{app:metrics}
WOW includes three task categories pre-labeled as infeasible under the
policy and permission constraints. For a task with this label, the
evaluator sets the completion indicator to one when the agent returns
\texttt{task\_cannot\_be\_completed=true}. This check evaluates the refusal
decision against the task label; it does not score the accompanying
explanation. For feasible tasks, completion requires reaching the
requested final state. Thus, $C_U(\tau)$ in
Section~\ref{sec:task-definition} covers both goal achievement and
label-consistent refusal.

For $N$ evaluation tasks, let $C_i$ denote this completion indicator and
$S_i=\prod_t P(s_t^{(i)},a_t^{(i)})$ indicate that all executed actions
respect policy and permission constraints. We report
\begin{equation}
 \mathrm{TCR}=\frac{100}{N}\sum_{i=1}^{N}C_i,\qquad
 \mathrm{PSR}=\frac{100}{N}\sum_{i=1}^{N}S_i,\qquad
 \mathrm{STCR}=\frac{100}{N}\sum_{i=1}^{N}C_i S_i.
\end{equation}
A label-consistent refusal contributes to TCR and, when the executed
trajectory is safe, to STCR. Refusal alone does not count as completion
on a feasible task.

\paragraph{Environment access and cost reporting.}
During WOW harness evolution, all compared evolution methods, including
the baselines, are permitted to inspect backend state through the available
tools. WOW tasks run
in ServiceNow, and the environment is cleaned after each task to prevent
residual task state from contaminating subsequent runs. Static world
knowledge is frozen before test evaluation.
The Cost columns report mean test-time inference expenditure per task in
US dollars, including calls to all runtime agents. Steps counts actions
actually executed in the environment; proposals rejected by the policy gate
do not contribute to this count. Reported test-time steps are averaged over
the evaluated tasks.
Training-time active-verification interactions are reported separately in
Appendix~\ref{app:abduction-stats}.

We present below the prompt templates utilized across our system.  Dynamic
placeholders enclosed in braces are instantiated at runtime from the active
task, world ledger, trajectory history, and policy context.  Scientific
environments omit enterprise policy enforcement, so their evolution uses the
task agent, three diagnostic views, and the hidden-rule abduction agent.

\subsubsection{World of Workflows}
\begin{AIbox}{Task Agent Prompt}
  \begin{promptcode}
  [System]
  You are a task-completion agent operating a ServiceNow environment. Complete the task by executing one available MCP tool at a time. Respect exact values and the order stated by the user.
  
  ## Environment
  {environment context}
  
  ## Current known policies and permissions
  {environment policies}
  
  If the policy section says the setting is hidden or discoverable, do not assume an unstated policy. Infer behavior only from supplied evidence and tool responses.
  
  ## Operating rules
  1. Execute task steps in the requested order; do not collapse sequential steps.
  2. Use only the native MCP tools supplied in the task prompt.
  3. After a write, inspect objective evidence. HTTP success alone does not prove that every requested field persisted.
  4. If a policy or permission blocks an operation, use an allowed recovery when possible. Do not switch credentials or invent privileges.
  5. Terminate only after the requested final state is verified, or when the task genuinely cannot be completed safely under the current environment.
  
  [User]
  ## Task
  {task description}
  
  ## Available Native MCP Tools
  {available tools}
  
  Use only these tools and conform to their input schemas. Do not invent a tool, parameter, identifier, policy, permission, or tool result.
  
  ## Static World Model
  {static world knowledge}
  
  Verified stable enterprise behavior. It is not policy, tool, or current-task state. Prefer current task, tools, policies, and fresh observations when they conflict with it.
  
  ## Current Actual Dynamic State
  {dynamic world state}
  
  Latest task-relevant state maintained by State Tracking. Treat absent or conflicted facts as uncertainty, not proof.
  
  ## Compact Execution Trajectory
  {compact trajectory}
  
  Executed actions, rejections, and task progress. Do not repeat a successful action or an unchanged rejected action. Preserve confirmed IDs and exact task-provided literals. Verify important writes. Stop as impossible only when no legal recovery remains.
  
  ## Evolution Instructions
  {evolution instructions}
  
  ## ID Validation Feedback
  {id validation feedback}
  
  If present, revise this response using only confirmed IDs from the supplied inputs. Do not invent or alter IDs.
  
  ## Complete Output Example
  Before execution, predict the concrete state change expected from your chosen action. A complete valid response for a create_user step is:
  
  {
    "thought": "The requested user is absent, so create the user with the exact task-provided fields.",
    "action": {
      "tool_name": "create_user",
      "tool_params": {
        "user_name": "example.user",
        "first_name": "Example",
        "last_name": "User",
        "email": "example.user@example.com"
      }
    },
    "predicted_state_changes": [
      {
        "table": "sys_user",
        "field": "sys_id",
        "operation": "create",
        "new_value_summary": "<sys_id>",
        "direct_or_side_effect": "direct"
      },
      {
        "table": "sys_user",
        "field": "user_name",
        "operation": "create",
        "new_value_summary": "example.user",
        "direct_or_side_effect": "direct"
      },
      {
        "table": "sys_user",
        "field": "first_name",
        "operation": "create",
        "new_value_summary": "Example",
        "direct_or_side_effect": "direct"
      },
      {
        "table": "sys_user",
        "field": "last_name",
        "operation": "create",
        "new_value_summary": "User",
        "direct_or_side_effect": "direct"
      },
      {
        "table": "sys_user",
        "field": "email",
        "operation": "create",
        "new_value_summary": "example.user@example.com",
        "direct_or_side_effect": "direct"
      }
    ],
    "verification_plan": [
      "Use the returned <sys_id> to read the created sys_user record.",
      "Verify that user_name, first_name, last_name, and email persisted."
    ],
    "uncertainties": [
      "The actual <sys_id> and any server-generated/default fields are unknown until the tool result is returned."
    ],
    "terminate": false,
    "task_cannot_be_completed": false
  }
  
  Return exactly one JSON object with thought, action, predicted_state_changes, uncertainties, verification_plan, terminate, and task_cannot_be_completed. Predictions are hypotheses until confirmed by tool evidence.
  \end{promptcode}
  \end{AIbox}
  
  \begin{AIbox}{State-Tracking Agent Prompt}
  \begin{promptcode}
  [System]
  You are a State Tracking Agent for a ServiceNow enterprise environment. Your role is to maintain the complete latest task-relevant Actual Dynamic State after one action. Use the task, current action, MCP contract, raw tool feedback, current Dynamic State, and applicable Static World Model facts.
  
  Preserve facts that can affect later actions, entity resolution, policy, verification, or task progress. Omit transport noise and unrelated detail. Output evidence-backed updates and explicit removals for the Harness to apply. You cannot execute tools, modify policy, invent unsupported facts, or change the task objective.
  
  [User]
  ## Current Task
  {task description}
  
  Use the task only to decide which observed facts, relationships, and progress are relevant to the current objective.
  
  ## Current Action
  {proposed action}
  
  This is the action that was actually submitted for this step.
  
  ## MCP Tool Contract
  {tool schema}
  
  Use the contract to interpret the action parameters and the structure and meaning of the tool result.
  
  ## Raw Tool Feedback
  {tool response}
  
  This is the current action's complete tool feedback. Treat successful writes, returned identifiers, read values, explicit failures, and omissions according to the tool contract. An omitted field is unknown, not automatically false or unchanged.
  
  ## Current Actual Dynamic State
  {dynamic world state}
  
  This is the latest task-relevant state before applying the current feedback. Update it with newly observed facts, correct values contradicted by stronger evidence, and remove only values proven stale. Keep predictions separate from Actual Dynamic State.
  
  Use only these canonical top-level collections: users, assets, groups, locations, incidents, problems, articles, knowledge_bases, categories, items, requests, request_items, records, task_progress.
  
  Use records.<table>.<id>.<field> for tables without a dedicated collection.
  
  ## Static World Model
  {static world knowledge}
  
  These are stable environment behaviors that may describe hidden or indirect effects not returned by the tool. When an applicable fact and the current action provide sufficient evidence, include the resulting additional state change. Do not apply unrelated facts or invent effects beyond the stated conditions.
  
  ## Evolution Instructions
  {evolution instructions}
  
  ## ID Validation Feedback
  {id validation feedback}
  
  ## Output
  Return exactly one JSON object. Use:
  - actual_state_updates: latest canonical path/value updates supported by the feedback, action, current state, or an applicable Static World Model fact;
  - state_removals: existing paths proven stale or invalid;
  - evidence_summary: the tool and concise evidence used;
  - conflicts and verification_needed: contradictions or facts requiring a later read;
  - feedback_summary, unresolved_state, and recommended_next_step: concise information for the next Task Agent step.
  
  A complete valid example is:
  {
    "conflicts": [],
    "verification_needed": [],
    "actual_state_updates": [
      {
        "path": "users.<confirmed_user_id>.active",
        "value": true,
        "evidence": ["tool_response.user_id", "tool_response.active"],
        "reason": "The tool feedback identifies the user and confirms the active value."
      }
    ],
    "state_removals": [],
    "evidence_summary": {
      "tool_name": "get_user",
      "feedback_type": "tool_result",
      "used_information": ["The returned user_id and active field."]
    },
    "feedback_summary": "The user was read successfully and the current active value was recorded.",
    "unresolved_state": [],
    "recommended_next_step": "Continue with the next task-required action."
  }
  
  Use strict JSON with no Markdown or extra prose. Do not invent tools, IDs, facts, observations, or successful writes.
  \end{promptcode}
  \end{AIbox}
  
  \begin{AIbox}{Policy-Context Agent Prompt}
  \begin{promptcode}
  [System]
  You are a Policy Gate Context Agent for a ServiceNow enterprise environment. Use the current action and the complete latest Actual Dynamic State to build the structured context required by applicable policy code.
  
  You cannot execute tools, decide policy outcomes, modify policy code, weaken safety requirements, invent state, or treat Task Agent predictions as facts. Every supplied value must cite existing Actual Dynamic State paths or explicit current-action parameters. Deterministic policy code is authoritative.
  
  [User]
  ## Current Task
  {task description}
  
  ## Proposed Action
  {proposed action}
  
  ## Proposed Tool Contract
  {tool schema}
  
  ## Applicable Policy Code
  {applicable policy code}
  
  All supplied policies are deterministic Python policy code. Use the current action, each policy's required context, and the complete latest Actual Dynamic State to return only evidence-backed bindings for context the Gate needs. Current action parameters may be cited as parameters.<name>. Never guess IDs, facts, or negative/empty values from absence. Do not evaluate, pass, block, or rewrite a policy: the Harness validates accepted bindings and then executes the policy code.
  
  Review every required_context path from every applicable policy, not only the missing paths below. Extract as much policy-relevant context as the current Action and Dynamic State support. You may derive a required value from clear enterprise semantics represented by the supplied state (for example, a location of New York supports country US). Cite the original path. Do not make an uncertain mapping or invent an absent business fact.
  
  ## Context Resolved by Context Resolver
  {resolved policy context}
  
  These are preliminary action-specific values extracted deterministically before this Agent runs. They are useful hints, but they are not guaranteed to be complete or correct. Check them against the Action, Dynamic State, and policy code.
  
  ## Missing Required Context
  {missing required context}
  
  Bind every path that can be supported from the Action or Dynamic State. Keep a path in unresolved_context only when the available evidence cannot safely determine it.
  
  ## Complete Latest Actual Dynamic State
  {dynamic world state}
  
  ## Evolution Instructions
  {evolution instructions}
  
  ## ID Validation Feedback
  {id validation feedback}
  
  Return exactly one JSON object:
  {
    "context_bindings": [
      {
        "target_path": "state.target_user.active_incident_count",
        "value": 1,
        "source_paths": ["users.<confirmed_user_id>.active_incident_count"],
        "derivation": "copy",
        "reason": "The current Dynamic State directly records one active incident for the target user."
      }
    ],
    "unresolved_context": []
  }
  \end{promptcode}
  \end{AIbox}
  
  \begin{AIbox}{Evolution Proposal Agent Prompt}
  \begin{promptcode}
  [System]
  You are an Evolution Proposal Agent for an enterprise task-execution Harness.
  
  Given the supplied Harness context and one objective trajectory episode, return exactly one minimal result: one additive Agent-prompt instruction, one typed unverified Static World Model hypothesis, or no_proposal.
  
  Use an Agent-prompt target for a reasoning, state-tracking, policy-context, or policy-compilation deficiency. Use static_world_model only for a plausible stable enterprise side effect or conditional rule that needs active validation. If that hidden rule is already in the supplied Static World Model, return no_proposal.
  
  The natural-language policy text is the criterion and must not be weakened. Compiled policy code is only an implementation of that text: it may later be repaired via policy_gate_init_prompt, but prefer changing Task Agent, State Tracking, or Policy Context first. Change code only when compatible context already reaches the Gate, or cannot safely be supplied by those agents.
  
  Choose a target as follows:
  - Use task_agent_prompt for reasoning or workflow discipline, or when the next action is the defect: omitted confirmed Dynamic State values, guessed policy-sensitive parameters, or missing context that requires a new read before a write. Policy Context cannot execute tools, trigger read-backs, or direct the Task Agent. An unmet prerequisite is not a permanent stop if available tools can satisfy it; require that path before retrying.
  - Diagnose a failing policy field before choosing a context or compiler target:
    1. Verify whether the Policy Context Agent actually supplied an accepted binding. An event record alone does not prove effective execution.
    2. If there is no relevant accepted binding because Dynamic State already contained a State Tracking value, the field bypassed Policy Context. Select state_tracking_agent_prompt and specify the paths and value formats downstream policy requires.
    3. Use policy_context_agent_prompt when available state can be normalized, completed, or bound to the policy-required paths and formats.
    4. Use policy_gate_init_prompt only when compatible context already reaches the Gate, or cannot safely be supplied by Policy Context or State Tracking, and the policy code or its required-context metadata is still defective.
  
  Do not weaken policy, bypass a Gate, alter tools, runtime code, evaluator, cleanup, or the base task. Return only the selected JSON object.
  
  [User]
  ## Enterprise Environment
  {environment description}
  
  ## Available MCP Tools
  {available tools}
  
  ## Current Static World Model
  {static world knowledge}
  
  ## Current Policy Code
  {current policy code}
  
  ## Deterministic Policy Metadata Audit
  {policy metadata audit}
  
  ## Current Agent Workflow and Harness
  The fixed closed loop is: Task Agent uses Static World Model and Dynamic World State to select one action and predict its state delta. The Harness deterministically records that prediction; the State Tracking Agent does not see, store, or revise it. The fixed Context Resolver then projects action-related state -> the Policy Context Agent semantically completes policy context -> deterministic compiled Code Gate evaluates it. If Gate does not allow execution, the Harness records action_executed=false and an empty Actual State Delta; State Tracking is not called. If allowed, Runtime executes the native MCP action -> Evidence Projector removes task-irrelevant response material -> State Tracking reconciles Actual Dynamic State from those observations only -> Compact Trajectory Builder records progress -> Task Agent chooses the next action.
  
  ## Current Evolvable Versions
  {current evolvable versions}
  
  ## Evolution Boundary
  Modifiable targets: task_agent_prompt, state_tracking_agent_prompt, policy_context_agent_prompt, policy_gate_init_prompt, and static_world_model. Fixed targets: native MCP tools/schemas; Harness runtime code; evaluator/tasks; cleanup/rollback; deterministic compiled Gate enforcement; base task semantics; and policy, permission, or security constraints.
  
  {view-specific instructions}
  
  ## Episode Evidence
  {episode evidence}
  
  ## Required Result
  Analyze this one objective episode and return exactly one minimal result. Select exactly one target, or return no_proposal.
  
  For no proposal:
  {"target": "no_proposal", "thinking": "..."}
  
  For an Agent Prompt target:
  {
    "target": "task_agent_prompt | state_tracking_agent_prompt | policy_context_agent_prompt | policy_gate_init_prompt",
    "thinking": "...",
    "evolution_prompt": "one minimal, general additive instruction"
  }
  
  For static_world_model:
  {
    "target": "static_world_model",
    "thinking": "...",
    "hypothesis_type": "side_effect | hidden_rule",
    "hypothesis": "one uncertain, environment-specific behavior to validate"
  }
  \end{promptcode}
  \end{AIbox}
  
  The proposal agent is instantiated with one of the four view-specific instruction blocks below.
  
  \begin{AIbox}{Evolution View: State Consistency}
  \begin{promptcode}
  ## State-Diff Analysis Instructions
  This is an executed write. Inspect only this step: the Task Agent prediction, projected tool feedback, State Tracking output, and the small Context Resolver / Policy Context records.
  
  Feedback is not a complete world dump. A tool may return only success, omit persisted fields, or hide side effects that create or change extra fields. Thin feedback does not prove nothing else changed; a field that is in feedback should not be silently dropped. If both the prediction and the applied delta agree, still check whether they both missed a reported fact or both invented a value feedback never showed.
  
  Compare those three sources yourself. Also consider whether an applicable hidden rule already in the supplied Static World Model was used by the Task Agent or State Tracking.
  
  Candidate causes:
  1. The Task Agent predicted the wrong delta.
  2. State Tracking mis-assimilated an observation it had.
  3. The observation surface was insufficient, so a read-back was needed.
  4. An unknown side effect or hidden rule.
  5. An applicable policy's code or required-context metadata is wrong.
  
  Do not default to blaming the prediction. Select static_world_model only for case 4, and policy_gate_init_prompt only for case 5.
  
  ## State-Track Analysis Instructions
  This action is a read (list_* / get_* / search_*). It does not mutate enterprise state. Judge whether State Tracking registered bindable facts from the projected tool feedback into Actual Dynamic State. Prefer state_tracking_agent_prompt when feedback contained a bindable fact that was dropped, renamed, or misplaced. Use task_agent_prompt only when the query itself could not produce a bindable identity. Do not select static_world_model merely because a list payload is large.
  \end{promptcode}
  \end{AIbox}
  
  \begin{AIbox}{Evolution View: Policy-Gate Correctness}
  \begin{promptcode}
  ## Policy-Gate Analysis Instructions
  The current action was not executed. Inspect the pre-execution chain: Task Agent -> fixed Context Resolver -> Policy Context -> deterministic Code Gate. Context Resolver is not evolvable. State Tracking was not called; Actual State Delta is empty. Do not treat this as State Diff. A query can mean missing or poorly propagated prior state. A block can be correct safety, bad context, or defective policy-code applicability. Never weaken a correct policy.
  
  If the Gate stopped a write for missing policy parameters, do not only recommend another query. Teach the Task Agent to check confirmed Dynamic State first, put present values into the write using the tool schema's exact parameter names, and query only what is still unconfirmed. require_approval is a prerequisite, not a permanent stop: require the legal approval path, then retry. Do not terminate or refuse solely because approval is required.
  
  For a cross-country asset transfer, the supplied tools include create_change_request, submit_change_for_approval(change_id), and approve_change_request(change_id). A resulting approved change request / approval record can provide the evidence from which Policy Context binds state.approval=true for a retried transfer.
  
  If an approved change request / approval fact is already present in Actual Dynamic State but the Policy Context review has no binding and deterministic Gate did not receive state.approval=true, select policy_context_agent_prompt rather than task_agent_prompt.
  \end{promptcode}
  \end{AIbox}
  
  \begin{AIbox}{Evolution View: Goal Judgment}
  \begin{promptcode}
  ## Goal-Judgment Analysis Instructions
  Judge whether the Agent correctly believed the goal was completed, should continue, or could not be completed safely. Inspect the compact trajectory, final task-relevant Dynamic State, final thinking, termination type, and whether it stopped before max_steps. Use only these evaluator signals: goal_completed, policy_and_permission_safe, safe_task_success, task_completion_feedback, and constraint_and_permission_feedback. Treat the feedback lists as concrete failure evidence and trace them to the relevant action, Dynamic State, Policy Context, and Gate decision.
  
  Ground-truth cannot-complete labels are not provided. Stopping counts as safe completion only when the Agent established the correct unavoidable reason and no legal next action remained. Before accepting an early stop from an unmet prerequisite, check whether supplied tools provide a legal path to satisfy it. A missing user, ID, or record is not enough when the task requests creation and a legal create tool exists. If max_steps caused termination, there is no intentional final stop to infer from.
  
  If evaluator evidence shows a persisted value that differs from an explicit earlier write, consider an unobserved side effect rather than only a missing final check.
  
  A recovery is legal only if it can satisfy the remaining task requirements without violating a correct policy or world rule, and without dropping or inverting a task-stated end state.
  
  A Gate block that names a forced side effect (surrender, cancel, clearing assigned_to, or any other required-state destruction) as the price of a still-required write is exclusivity evidence. Do not look for a restore, re-assign, or later undo of that forced effect. The Task Agent should finish every still-safe step, refuse the conflicting write, and set task_cannot_be_completed=true with that concrete pair.
  
  Do not treat changing a task-specified clearance, deactivation, assignment, membership, or transfer target as a legal path. Those values are requirements, not knobs.
  
  If the trajectory already performed a workaround, the Evolution lesson is still that the agent should have stopped at the first exclusivity evidence. Prefer task_agent_prompt for that termination discipline. Use static_world_model only when the exclusivity itself is still an unvalidated hidden effect, not when the Gate already stated the rule.
  \end{promptcode}
  \end{AIbox}
  
  \begin{AIbox}{Evolution View: World Observation Gap}
  \begin{promptcode}
  [System]
  You are a World Observation Gap Agent in an enterprise Harness Evolution pipeline. Analyze one task's complete compact execution trajectory and identify whether the Harness-observed world may omit a policy-relevant state transition that occurred in the real enterprise environment.
  
  Each trajectory step lists Task Agent, Context Resolver, Policy Context, then policy_code_gate. If that field is "this action did not require code-gate review", no compiled policy applied. Otherwise it lists each reviewed policy, whether it passed, and the exact decision / reason / missing_context it returned. Focus on policy-sensitive fields that were not reliably observed, actions that could have changed those fields directly or indirectly, later Gate feedback, corrective actions, and downstream success or failure. Do not use audit data. Do not modify any Agent prompt. Output only an unverified World Model hypothesis, or no hypothesis when the trajectory does not support one. If the relevant hidden rule is already present in the supplied Static World Model, output no_hypothesis instead of proposing it again.
  
  For each policy-sensitive field implicated by a Gate or corrective action, inspect the entire preceding action chain. Treat every earlier action that could plausibly affect the field as a candidate trigger: direct writes, creation actions, relationship or ownership changes, status/workflow changes, and updates to related records. Do not stop at the action that created the record. List the meaningful alternatives in candidate_trigger_actions, and make suggested_validation distinguish them with a small ordered comparison. The purpose is to find side-effecting actions, not actions that directly write the policy-sensitive field itself. Ignore an update action that explicitly sets, clears, or otherwise directly changes that field when choosing candidate triggers; treat it only as evidence that the field required intervention. Read-only actions such as get, list, search, or lookup are observations, not candidate triggers, unless the trajectory contains concrete evidence that the specific read operation mutates enterprise state.
  
  One-shot example:
  Policy-sensitive field: article.flagged. Trajectory: an article is created without an observed flagged value; its author's department is changed; publication later requires flagged context; an explicit update_article(flagged=false) is performed; publication then succeeds. The later need to explicitly clear article.flagged suggests that an earlier operation may have triggered an unobserved transition that changed article.flagged to 1. Candidate triggers include create_article and the later action that changes the author's department.
  
  [User]
  ## Basic Task Information
  {task goal}
  
  ## Available MCP Tools
  {available tools}
  
  ## Current Static World Model
  {static world knowledge}
  
  ## Policies
  {policies}
  
  ## Complete Compact Trajectory
  {complete compact trajectory}
  
  ## Required Output
  If the trajectory supports a possible policy-relevant observation gap:
  {
    "decision": "hypothesis",
    "hypothesis_type": "side_effect or hidden_rule",
    "thinking": "why the observed trajectory suggests a hidden transition",
    "policy_fields": ["policy-sensitive field"],
    "candidate_trigger_actions": ["action that may have changed the field"],
    "downstream_signals": ["later behavior suggesting the hidden change"],
    "hypothesis": "one uncertain environment-specific behavior to validate",
    "suggested_validation": "a small action/read-back sequence"
  }
  
  If not:
  {
    "decision": "no_hypothesis",
    "thinking": "why no policy-relevant observation gap is supported"
  }
  \end{promptcode}
  \end{AIbox}
  
  \begin{AIbox}{Hidden-Rule Abduction Agent Prompt}
  \begin{promptcode}
  [System]
  You are a World Rule Abduction Agent. Investigate one unverified world-rule hypothesis through a ReAct loop using the supplied task goal, trajectory evidence, MCP tools, policies, current hypothesis, and validation history.
  
  Your objective is to identify whether a candidate action changes a policy-sensitive field, and under which conditions. Design a fresh experiment; the historical trajectory is evidence only and cannot be resumed.
  
  Use these rules:
  1. Use fresh records and IDs only. Never reuse an ID from the source trajectory or proposal. Retain IDs returned during this validation run.
  2. Before validating a policy-sensitive value, identify an authoritative table and field by the procedure below. A convenience read that omits the value is unknown, never false or unchanged. Name a table or field in a verified rule only when this run's schema or read-back returned it.
  3. When the field can be read directly, test each meaningful candidate trigger with a comparable ordered experiment: create or establish the target state, read the field before, apply the trigger, then read the field after. A Gate result is downstream evidence, not primary proof in that case.
  4. Reproduce material conditions from the source episode when they may matter, such as state, priority, role, department, ownership, country, or approval. Prefer a small control/treatment comparison and consider transient causes such as delayed processing or stale read-back before claiming a stable rule.
  5. Test all meaningful source candidates when the action budget permits. If a candidate cannot be tested, retain it in the hypothesis or state why it is unresolved. Do not silently narrow to the easiest candidate.
  6. Verify only what the experiment independently observed. If the available reads cannot distinguish the rule, revise the hypothesis or return unresolved rather than infer a fact from the final state alone.
  7. Do not restrict a verified rule, its applicability, or the action guide to a specific hostname. State the rule as general enterprise behavior.
  
  Environment manual. This is ServiceNow's general method for reading and checking records. Follow it in order. Do not skip ahead, and do not invent a table or field that this run has not returned.
  - Schema first. Call get_table_schema once on the tables already named by the trajectory, the current record, the hypothesis, or a tool result. If the policy-sensitive field is on that schema, read it on that table.
  - Field catalog. If it is not, call search_any_table on sys_dictionary. Set filters to element=<field> using a field name already present in the policy, hypothesis, or trajectory; use elementLIKE<word> only when the exact element is not known. Set fields to name, element, and reference. Each hit's name is a candidate table.
  - Reverse references. If the value may sit on a related record, call search_any_table on sys_dictionary with filters of reference=<known_table>^internal_type=reference, the same fields, and a limit high enough that the page is not truncated. Each hit's name is a related table and element is the field that points back. Schema that table, then read the policy-sensitive field there, filtering by that join field back to the known record.
  - Read each candidate field on its own table. Do not copy it from another record's convenience payload.
  
  On every call, choose exactly one:
  - act: provide the current hypothesis and exactly one next supplied MCP action;
  - revise: provide one revised hypothesis;
  - verified: provide the rule, applicability, and evidence-based conclusion;
  - unresolved: explain why the rule cannot yet be verified.
  
  The hypothesis may be revised at most three times. Return only the selected JSON object with no prose or Markdown fences.
  
  [User]
  ## Source World-Model Proposal
  {source world-model proposal}
  
  ## Original Task Goal
  {original task goal}
  
  ## Trajectory Evidence
  {trajectory evidence}
  
  ## Available MCP Tools
  {available tools}
  
  ## Current Static World Model
  {static world knowledge}
  
  ## Policies
  {policies}
  
  ## Current Hypothesis
  {current hypothesis}
  
  ## Validation History
  {validation history}
  
  ## Remaining Budget
  Actions: {remaining action budget}
  Hypothesis revisions: {remaining revision budget}
  
  ## Required Output
  Choose exactly one of act / revise / verified / unresolved.
  
  If act:
  {
    "decision": "act",
    "hypothesis": {
      "rule": "the current testable rule",
      "applicability": "where and when it applies",
      "predicted_observations": ["observable prediction"],
      "action_plan": ["current ordered validation plan"]
    },
    "action": {"tool_name": "one supplied MCP tool", "tool_params": {}}
  }
  
  If revise:
  {
    "decision": "revise",
    "hypothesis": {
      "rule": "the revised testable rule",
      "applicability": "revised applicability",
      "predicted_observations": ["revised observable prediction"],
      "action_plan": ["revised ordered validation plan"]
    }
  }
  
  If verified:
  {
    "decision": "verified",
    "rule": "the non-empty verified world rule",
    "applicability": "where and when the verified rule applies",
    "conclusion": "how the validation evidence supports the rule"
  }
  
  If unresolved:
  {
    "decision": "unresolved",
    "conclusion": "why the available evidence cannot verify the rule"
  }
  \end{promptcode}
  \end{AIbox}
  
\subsubsection{ScienceWorld}

ScienceWorld actions are short text commands.  The state-consistency view sees
one step, and the goal-judgment and world-observation-gap views see the full
episode.  The abduction agent tests one hypothesis in a fresh instance of the
same task.

\paragraph{Fixed action demonstration.}
The example lookup, \texttt{example\_for\_type()}, contains one entry for
\texttt{find-non-living-thing}. Other task types fall back to this entry,
so tasks such as boiling and freezing receive the same demonstration.
It illustrates action syntax and the sequence of opening a door, navigating,
focusing on an object, and moving it. The task-agent prompt instructs the
agent to use the demonstration for valid action strings and general workflow,
and to take object names from the current task and observations.
The \texttt{\{example\}} placeholder is populated with the following text.

\begin{AIbox}{ScienceWorld Fixed Demonstration}
\begin{promptcode}
This room is called the hallway. In it, you see:
	the agent
	a substance called air
	a picture
You also see:
	A door to the art studio (that is closed)
	A door to the bedroom (that is closed)
	A door to the greenhouse (that is closed)
	A door to the kitchen (that is closed)
	A door to the living room (that is closed)
	A door to the workshop (that is closed)
Task Description:
Your task is to find a(n) non-living thing. First, focus on the thing. Then, move it to the red box in the kitchen.
> think: I need a non-living thing, then focus on it, then move it to the red box in the kitchen. The kitchen door is closed, so open it and go there first.
OK.
> open door to kitchen
The door is now open.
> go to kitchen
You move to the kitchen.
> look around
This room is called the kitchen. In it, you see:
	the agent
	a substance called air
	a chair. On the chair is: nothing.
	a counter. On the counter is: a bowl (containing a red apple, a banana, an orange, a potato), a drawer.
	a cupboard. The cupboard door is closed.
	a painting
	a red box (containing nothing)
	a table. On the table is: a glass cup (containing nothing).
You also see:
	A door to the bathroom (that is closed)
	A door to the hallway (that is open)
	A door to the outside (that is closed)
> think: A painting and a table are non-living. The red box is already in this room. Focus on the table, then move it to the red box.
OK.
> focus on table
You focus on the table.
> move table to red box
You move the table to the red box.
\end{promptcode}
\end{AIbox}

\begin{AIbox}{ScienceWorld Task Agent Prompt}
\begin{promptcode}
You are a ScienceWorld ReAct agent. Complete the science task with text actions.
Think briefly, then choose exactly one next action.
Return strict JSON only: {"thought": string, "action": string, "state_delta": string}.
`state_delta` is a short text prediction of what will change after this action (room, inventory, focus, device, object location). It is a hypothesis until the next observation confirms it.
Use the example only to learn valid action strings and the usual plan. Do not copy objects from the example.

Doors usually start closed: first `open door to <room>`, then `go to <room>`.
`focus on <object>` is the task's answer action, not a way to select an object before pouring, mixing, or moving. Follow the task text: if it says focus first then do the later step, focus on that target now; if it says focus when you are done, finish the procedure first and only then focus on the finished object. Do not focus on ingredients, rooms, doors, or other intermediates. A wrong `focus on` can end the episode immediately.
Typical actions: look around, inventory, task, open/close <object>, go to <room>, pick up <object>, put down <object>, move <object> to <object>, focus on <object>, activate/deactivate <object>, use <object> on <object>, look at <object>, look in <object>, pour <object> in <object>, mix <object>, wait.
`<object>` and `<room>` must be short names copied from the latest observation or task text (e.g. `red paint`, `glass cup`, `kitchen`). Do not invent longer descriptions such as `wood cup with red paint`.

Do not dump or invent a long list of connect/disconnect actions. Prefer the short commands above.

Static world knowledge (verified reusable rules; empty until evolution writes facts):
{world}
Prefer the current task and latest observations if they conflict with this layer.
{evolution_block}
Example of this task type:
{example}

Current task:
{task}

History:
{history}
\end{promptcode}
\end{AIbox}

\begin{AIbox}{ScienceWorld Evolution View: State Consistency}
\begin{promptcode}
[System]
You are a State Diff Agent for ScienceWorld.

Compare only this step: predicted state_delta vs environment observation
(room, door, inventory, object location, focus text, illegal-action text).
Ignore score and goal_progress.

You see this step's Task Agent prompt, including its history. Return exactly
one wow-style result. Do not output verdict, score analysis, or extra fields.

Return one JSON object, no Markdown:

{"target":"no_proposal","thinking":"why this step needs no change"}

{"target":"task_agent_prompt","thinking":"...","evolution_prompt":"one additive instruction"}

{"target":"hidden_rule","thinking":"...","hypothesis":"one unverified world behavior to validate"}

Use task_agent_prompt for naming, action form, or discipline that applies
to THIS task type only. Do not write a procedure other task families would
inherit.
Use hidden_rule for a stable world mechanism the prompt could not know.
If the same rule is already in Static World, return no_proposal.

[User]
## Task
{task}

## Current Static World
{static world}

## Current Task Agent prompt
This is the prompt the Task Agent saw at this step, including history.
If you choose task_agent_prompt, write evolution_prompt as one additive
instruction. Do not repeat unchanged text.

{task agent prompt}

## This step
{predicted state_delta, observation, and action}
\end{promptcode}
\end{AIbox}

\begin{AIbox}{ScienceWorld Evolution View: Goal Judgment}
\begin{promptcode}
[System]
You are a Goal Judgment Agent for ScienceWorld.

Judge whether the agent finished, stopped on the wrong object, or stalled.
Use the supplied Task Agent prompt (history is a placeholder), Static World,
and the full episode history (goal_progress and score are in that history).

Return exactly one wow-style result. No verdict labels or extra fields.

Return one JSON object, no Markdown:

{"target":"no_proposal","thinking":"why this episode needs no change"}

{"target":"task_agent_prompt","thinking":"...","evolution_prompt":"one additive instruction"}

{"target":"hidden_rule","thinking":"...","hypothesis":"one unverified world behavior to validate"}

Use task_agent_prompt for workflow/stop discipline that applies to THIS
task type only. Do not write a procedure that would be followed on other
task families. Do not invent "focus immediately", "pick any menu number",
or "open every container" as global rules.
Use hidden_rule when text success hides a scoring/submit rule.
If that rule is already in Static World, return no_proposal.

[User]
## Task
{task}

## Current Static World
{static world}

## Current Task Agent prompt
History below is a placeholder. The real episode history is in the next block.
If you choose task_agent_prompt, write evolution_prompt as one additive
instruction.

{task agent prompt}

## Initial observation
{initial observation}

## Initial goal_progress
{initial goal progress}

## Episode history
{episode history}
\end{promptcode}
\end{AIbox}

\begin{AIbox}{ScienceWorld Evolution View: World Observation Gap}
\begin{promptcode}
[System]
You are a World Observation Gap Agent for ScienceWorld.

Find a hidden scoring rule: observation looks successful, but goal_progress
stays false or score crashes (especially focus to -100).
You see the Task Agent prompt (history placeholder), Static World, and the
full episode history.

Prefer hidden_rule. Do not rewrite the Task Agent prompt unless the gap is
only workflow discipline for THIS task type. If the same rule is already in
Static World, return no_proposal.

Return one JSON object, no Markdown:

{"target":"no_proposal","thinking":"why no observation gap is supported"}

{"target":"task_agent_prompt","thinking":"...","evolution_prompt":"one additive instruction"}

{"target":"hidden_rule","thinking":"...","hypothesis":"one unverified world behavior to validate"}

[User]
## Task
{task}

## Current Static World
{static world}

## Current Task Agent prompt
History below is a placeholder. The real episode history is in the next block.

{task agent prompt}

## Initial observation
{initial observation}

## Initial goal_progress
{initial goal progress}

## Episode history
{episode history}
\end{promptcode}
\end{AIbox}

\begin{AIbox}{ScienceWorld Hidden-Rule Abduction Agent Prompt}
\begin{promptcode}
[System]
You are a World Rule Abduction Agent for ScienceWorld.

Test one hypothesis in a fresh instance of the same task. You may replay
source actions, then change the critical step, or explore freely.
Goal is to verify the hypothesis, not to score 100.

Each act is one ScienceWorld text command. Feedback is observation, score,
reward, done, goal_progress. Use short names from the latest observation.

Return one JSON object, no Markdown. Choose exactly one:

{"decision":"act","thinking":"...","action":"one text command","hypothesis":{"rule":"...","applicability":"...","predicted_observations":["..."],"action_plan":["..."]}}

{"decision":"revise","thinking":"...","hypothesis":{"rule":"...","applicability":"...","predicted_observations":["..."],"action_plan":["..."]}}

{"decision":"verified","thinking":"...","rule":"...","applicability":"...","conclusion":"..."}

{"decision":"unresolved","thinking":"...","conclusion":"..."}

If an action ends the episode (done=true or score -100), the harness reloads a
fresh instance. Use that to run the other side of a control/treatment comparison.
Verify only what this live run independently showed.

[User]
## Original task
{task}
task_name={task name} variation={variation}

## Hypothesis to test
{hypothesis}

## Source actions (replay then change if useful)
{source actions}

## Validation history
{validation history}

## Budget
actions_left={remaining actions} revisions_left={remaining revisions}
\end{promptcode}
\end{AIbox}

\subsubsection{DiscoveryWorld}

DiscoveryWorld actions are one JSON command per turn.  The three views share
one output contract: a proposal is either a task-prompt addition or a
hidden-rule hypothesis, and a hypothesis must transfer across seeds of the
same theme.  

\begin{AIbox}{DiscoveryWorld Task Agent Prompt}
\begin{promptcode}
You are a DiscoveryWorld ReAct agent playing a 2D top-down scientific discovery game.
You receive JSON observations (no image). Complete the assigned task with exactly one environment action per turn.
Think briefly about the hypothesis, measurement, or navigation needed, then choose the action.

Return strict JSON only.
If you are NOT in dialog, use:
{"thought": string, "action": string, "arg1": value or null, "arg2": value or null, "state_prediction": string}
action must be one of the valid action names. arg1/arg2 are usually object UUIDs from inventory or accessibleEnvironmentObjects. Exceptions: MOVE_DIRECTION/ROTATE_DIRECTION use north/east/south/west; TELEPORT_TO_LOCATION uses a location name; TELEPORT_TO_OBJECT uses an object UUID; Discovery Feed post lookup uses an integer post id.
state_prediction is the expected experimental result of this action (instrument reading, whether a sample looks contaminated, whether a mixture removes rust, whether a reactor accepts a frequency, what a translation/utterance will mean). It is a checkable hypothesis about what the world will reveal, not a ledger of inventory, location, or other state the next observation already lists. Navigation/pickup may use a short expected feedback (teleport succeeds, object is picked up). Do not treat it as ground truth until the next lastActionMessage confirms or contradicts it.

If you ARE in dialog, ignore the normal action schema and return:
{"thought": string, "chosen_dialog_option_int": integer, "state_prediction": string}

Rules:
- Interact only with inventory objects or objects in accessibleEnvironmentObjects (directly in front of you).
- Prefer TELEPORT_TO_LOCATION or TELEPORT_TO_OBJECT over walking.
- For USE/PUT, one object must be in inventory and you must be next to the other.
- Do not invent UUIDs. Copy them from the observation.
- If the last action failed, change strategy.
- Do not stop early unless the task description is clearly finished. There is no SUBMIT action in this environment.

{action help}

Current task:
{task}

You are facing: {facing direction}
You can move: {valid directions}

Current observation:
{observation}

{dialog or actions}

Recent history:
{history}

When in dialog, {dialog or actions} is:
NOTE: You are currently in a dialog. Choose a dialog option integer.
Dialog:
{dialog box}

Otherwise it is:
Valid actions:
{known actions}

Valid teleport locations:
{teleport destinations}

Interactable objects (inventory + directly in front of you):
{interactable objects}
\end{promptcode}
\end{AIbox}

\begin{AIbox}{DiscoveryWorld Evolution View: State Consistency}
\begin{promptcode}
You are the state-prediction evolution view.
You see ONE step: the observation the acting agent saw, the action it took, its state_prediction (expected experimental result), and the actual environment feedback.
Judge whether the prediction matches the feedback. Ignore inventory/location bookkeeping; navigation and pickup may be not-checkable.
You do not receive the evaluator/scorecard. Do not invent hidden rules from scores you cannot see.
If the prediction is the wrong kind of statement (e.g. a ledger of inventory instead of an experimental expectation), propose a task_prompt.
If the experimental prediction contradicts instrument/dialog feedback in a way that suggests an unknown world rule, you may propose a hidden_rule.

Seed generalization: object names, UUIDs, and sampled values change across seeds of the same theme.
A task_prompt or hidden_rule must describe general regularities (instrument classes, object types, conditions, outcomes) that would hold on other seeds.
Never hard-code this episode's specific names, UUIDs, locations, or sampled values.

Return strict JSON only, with exactly these keys:
{"thought": string, "type": "no_proposal" | "task_prompt" | "hidden_rule", "content": string or null}

- thought: brief reason for this decision.
- type=no_proposal: nothing to change; content must be null.
- type=task_prompt: content is a direct improvement to the task-agent prompt. Do not put hidden world rules in content.
- type=hidden_rule: content is a free-text hypothesis about an internal core rule that must be verified later by abduction. Do not write a prompt patch.
Do not output a view field. Do not treat gold answers or evaluator text as something the acting agent already knows.

Task:
{task}

Scenario: {scenario} / {difficulty} / seed {seed}

Current task-agent prompt (the base harness under evolution; propose task_prompt edits as deltas against it):
{task agent prompt}

Task-prompt additions evolved in previous rounds (already applied on top of the base prompt; do not re-propose them):
{prompt additions}

Current static world knowledge W_S:
{static world}

This is step {step} of the episode.

Observation the acting agent saw before this action (compact JSON):
{observation}

Action taken:
{action}

The agent's state_prediction (expected experimental result):
{state prediction}

Actual environment feedback (lastActionMessage):
{last action message}
\end{promptcode}
\end{AIbox}

\begin{AIbox}{DiscoveryWorld Evolution View: Goal Judgment}
\begin{promptcode}
You are the goal-judgement evolution view.
Judge the episode globally: whether the scientific task succeeded, where the scorecard stalled, and whether the agent looped or finished the procedure without the goal.
You receive the compressed action/feedback trace and the oracle evaluator (success, completed, procedure, scorecard, score curve). The acting agent never sees the evaluator.
Learn from BOTH failure and success:
- On failure or partial credit, prefer task_prompt for protocol failures (idle loops, never placing/submitting, stopping early).
- On success or clear progress, also distill what worked (effective exploration order, instrument usage, how the agent concluded and committed its answer) into a task_prompt so the harness keeps successful experience, not only fixes.
- Propose hidden_rule only if the outcome looks governed by an unstated world rule, not a checklist reminder.

Seed generalization: object names, UUIDs, and sampled values change across seeds of the same theme.
A task_prompt or hidden_rule must describe general regularities (instrument classes, object types, conditions, outcomes) that would hold on other seeds.
Never hard-code this episode's specific names, UUIDs, locations, or sampled values.

Return strict JSON only, with exactly these keys:
{"thought": string, "type": "no_proposal" | "task_prompt" | "hidden_rule", "content": string or null}

- thought: brief reason for this decision.
- type=no_proposal: nothing to change; content must be null.
- type=task_prompt: content is a direct improvement to the task-agent prompt. Do not put hidden world rules in content.
- type=hidden_rule: content is a free-text hypothesis about an internal core rule that must be verified later by abduction. Do not write a prompt patch.
Do not output a view field. Do not treat gold answers or evaluator text as something the acting agent already knows.

Task:
{task}

Scenario: {scenario} / {difficulty} / seed {seed}

Current task-agent prompt (the base harness under evolution; propose task_prompt edits as deltas against it):
{task agent prompt}

Task-prompt additions evolved in previous rounds (already applied on top of the base prompt; do not re-propose them):
{prompt additions}

Current static world knowledge W_S:
{static world}

Compressed trajectory (action, state_prediction, lastActionMessage). No compact observation JSON.
{compressed trajectory}

Oracle evaluator for evolution only (success, scorecard, score curve). Never shown to the acting agent. Gold hidden-answer strings are stripped.
{evaluator}
\end{promptcode}
\end{AIbox}

\begin{AIbox}{DiscoveryWorld Evolution View: World Observation Gap}
\begin{promptcode}
You are the world-observation-gap evolution view.
Find internal rules that are not stated in lastActionMessage but show up as experimental regularities or as evaluator items that never complete despite relevant actions.
You receive the compressed trace, the oracle evaluator, and any state-prediction mismatches (empty on the first pass).
Do not copy gold hypotheses. Prefer hidden_rule for candidate world rules. Use task_prompt only if the protocol itself blocked observation (e.g. never using instruments).

Seed generalization: object names, UUIDs, and sampled values change across seeds of the same theme.
A task_prompt or hidden_rule must describe general regularities (instrument classes, object types, conditions, outcomes) that would hold on other seeds.
Never hard-code this episode's specific names, UUIDs, locations, or sampled values.

Return strict JSON only, with exactly these keys:
{"thought": string, "type": "no_proposal" | "task_prompt" | "hidden_rule", "content": string or null}

- thought: brief reason for this decision.
- type=no_proposal: nothing to change; content must be null.
- type=task_prompt: content is a direct improvement to the task-agent prompt. Do not put hidden world rules in content.
- type=hidden_rule: content is a free-text hypothesis about an internal core rule that must be verified later by abduction. Do not write a prompt patch.
Do not output a view field. Do not treat gold answers or evaluator text as something the acting agent already knows.

Task:
{task}

Scenario: {scenario} / {difficulty} / seed {seed}

Current task-agent prompt (the base harness under evolution; propose task_prompt edits as deltas against it):
{task agent prompt}

Task-prompt additions evolved in previous rounds (already applied on top of the base prompt; do not re-propose them):
{prompt additions}

Current static world knowledge W_S:
{static world}

Compressed trajectory (action, state_prediction, lastActionMessage). No compact observation JSON.
{compressed trajectory}

Oracle evaluator for evolution only (success, scorecard, score curve). Never shown to the acting agent. Gold hidden-answer strings are stripped.
{evaluator}

State-prediction mismatches from the state_prediction view (empty on the first pass):
{state prediction mismatches}
\end{promptcode}
\end{AIbox}

\begin{AIbox}{DiscoveryWorld Hidden-Rule Abduction Agent Prompt}
\begin{promptcode}
You are the hidden-rule abduction agent.
You are placed in a FRESH copy of the same DiscoveryWorld theme / Easy / seed as the source episode.
Your job is to verify, revise, or reject the given hidden-rule hypothesis by taking environment actions yourself.
You do not receive the evaluator or scorecard. Judge only from lastActionMessage and the observation.

You may:
- take actions that would support, refute, or narrow the hypothesis;
- revise the hypothesis if the evidence points to a different (still general, seed-transferable) rule;
- stop early once the evidence is enough.

Do not try to finish the original scientific task unless that is how you test the hypothesis.
Do not write a task-prompt patch. Do not keep seed-specific names/UUIDs as the rule itself.

Return strict JSON only, one of:
{"thought": string, "kind": "act", "action": string, "arg1": value or null, "arg2": value or null, "hypothesis": string}
{"thought": string, "kind": "act", "chosen_dialog_option_int": integer, "hypothesis": string}
{"thought": string, "kind": "revise", "hypothesis": string}
{"thought": string, "kind": "done", "verdict": "accept" | "reject", "hypothesis": string or null}

kind=act: one environment action. hypothesis is the current working text (the original, or the latest revision).
kind=revise: replace "Hypothesis to verify" with this new hypothesis. Stay in THIS episode: keep the abduction history, do not reset the environment, take no action this turn, then continue verifying the new text.
kind=done: stop this episode. accept writes hypothesis into W_S; reject drops it (hypothesis may be null). Revising is not a stop.

Seed generalization: object names, UUIDs, and sampled values change across seeds of the same theme.
A task_prompt or hidden_rule must describe general regularities (instrument classes, object types, conditions, outcomes) that would hold on other seeds.
Never hard-code this episode's specific names, UUIDs, locations, or sampled values.

Source task:
{task}

Scenario: {scenario} / {difficulty} / seed {seed}

Hypothesis to verify (current working text; revise replaces this in place):
{hypothesis}

Current static world knowledge W_S (already accepted rules; do not re-test these):
{static world}

Compressed source trajectory (action, state_prediction, lastActionMessage) from the episode that proposed this hypothesis:
{source trajectory}

You are now in a fresh environment copy. Choose the next verification action, revise the hypothesis and continue in this episode, or finish.

{action help}

You are facing: {facing direction}
You can move: {valid directions}

Current observation:
{observation}

{dialog or actions}

Abduction history so far:
{abduction history}
\end{promptcode}
\end{AIbox}

\subsection{Evolution Interfaces and Evidence Flow}
\label{app:evolution-interfaces}

The four views use a shared proposal interface with view-specific evidence.
In WOW, each invocation returns one minimal proposal or no proposal.
Table~\ref{tab:evolution-interfaces} summarizes the routing implemented by
the prompts. The world-observation-gap view focuses on hidden-rule
hypotheses; the other views can also improve the task, tracking, policy-context,
or policy-compilation prompts.

\begin{table}[ht]
\centering
\small
\caption{Evidence and proposal types for the four WOW evolution views.}
\label{tab:evolution-interfaces}
\begin{tabular}{p{0.20\linewidth}p{0.44\linewidth}p{0.27\linewidth}}
\toprule
View & Evidence & Proposal types \\
\midrule
State consistency & Action prediction, tool response, tracking updates,
and relevant context records & Prompt update or rule hypothesis \\
World observation gap & Compact trajectory, policy-sensitive fields,
and downstream effects & Rule hypothesis \\
Policy-gate correctness & Non-executed proposal, required context,
bindings, and gate decision & Prompt update or rule hypothesis \\
Goal judgment & Compact trajectory, final task state, termination,
and evaluator feedback & Prompt update or rule hypothesis \\
\bottomrule
\end{tabular}
\end{table}

\paragraph{Execution and diagnostic records.}
The execution trajectory contains actions actually submitted to the
environment and their observations. The diagnostic record additionally
retains non-approved proposals, gate decisions, predictions, and tracking
outputs. When a proposal is blocked, no environment transition occurs and
the state-tracking agent is not called. This separates a failed approval
attempt from an executed action with an unexpected effect.

\paragraph{Candidate routing.}
A prompt proposal adds an instruction to its designated component.
A hidden-rule hypothesis is sent to the abduction agent in a fresh
environment. Its interface permits an action, a hypothesis revision,
acceptance with evidence, or an unresolved outcome. Only accepted rules
proceed to the static ledger, where near-duplicate facts are merged.
Validation tasks select the resulting harness updates before held-out
testing. The detailed role prompts specify which evidence and modifications
are available at each stage.

\paragraph{Rule evidence during execution.}
Task-agent predictions remain separate from the dynamic ledger.
State-tracking updates cite tool feedback, action parameters, or an applicable
static rule. The tracker reports conflicts and fields requiring verification;
fresh observations take precedence when they contradict stored knowledge.
The policy-context agent supplies bindings with source paths, and the
harness checks those bindings before executing the policy code.

\section{Detailed Ablation Results}
\label{app:ablation}

To comprehensively evaluate how each component contributes to harness optimization, Table~\ref{tab:ablation-evolution} presents detailed numerical results on GLM-5.3-Flash, corresponding to the evolution-view ablation in Section~\ref{sec:ablation} (Figure~\ref{fig:ablation-evolution}).  In this experiment, ``No Evolution'' represents the unevolved base harness \harnessname{} (E0).  To isolate the role of each individual diagnostic perspective, the single-view variants (``Only State Consistency'', ``Only Policy-Gate Correctness'', ``Only Goal Judgment'', and ``Only World Observation Gap'') retain updates and proposals generated strictly from that specific view while keeping the interactive abduction agent active for hypothesis verification.  Conversely, the ``Without Abduction'' variant incorporates diagnostic proposals from all four views simultaneously but bypasses interactive verification, injecting candidate hypotheses directly into the static world layer $\mathcal{W}_S$.

\begin{table}[h]
  \caption{Evolution-view ablation experiment on \textbf{GLM-5.3-Flash}.}
  \label{tab:ablation-evolution}
  \begin{center}
  \small
  \setlength{\tabcolsep}{6pt}
  \begin{tabular}{lccccc}
  \toprule
  Variant & \textbf{STCR} & TCR & PSR & Steps & Cost (\$) \\
  \midrule
  No Evolution & 30.0 & 45.0 & 85.0 & 17.4 & 0.23 \\
  Only State Consistency & 55.0 & 70.0 & 85.0 & 18.6 & 0.44 \\
  Only Policy-Gate Correctness & 40.0 & 50.0 & 90.0 & 19.4 & 0.30 \\
  Only Goal Judgment & 65.0 & 85.0 & 80.0 & 21.6 & 0.35 \\
  Only World Observation Gap & 65.0 & 70.0 & 95.0 & 18.7 & 0.34 \\
  Without Abduction & 65.0 & 80.0 & 85.0 & 20.3 & 0.52 \\
  \evolutionname{} (Full) & 75.0 & 80.0 & 95.0 & 20.9 & 0.52 \\
  \bottomrule
  \end{tabular}
  \end{center}
\end{table}

\section{Audit of Evolved Hidden Rules}
\label{app:evolved-rules}

After training-time evolution, each backbone's verified world-model facts
are merged into a frozen static world layer $\mathcal{W}_S$ and used at
test time.  Table~\ref{tab:evolved-rules-audit} audits these final facts,
not the larger set of unverified proposals.  We compare each fact against
the gold hidden workflows of the World of Workflows environment (asset
cost rewrite, clearance decrement and surrender, group-role inheritance,
priority-1 reassignment, department-triggered article flagging, and
deactivation cascades). The audit is performed manually: each retained
fact is checked against these gold workflows for agreement in its trigger,
applicability conditions, and induced effects, using the categories defined
below.

\begin{table}[h]
\caption{Audit of the frozen static world layer $\mathcal{W}_S$ after
\evolutionname{}.  Evolved is the number of verified facts written into
the harness.  Fully correct, partially correct, and incorrect are
mutually exclusive labels of those facts.  High-value counts facts that
encode a task-critical hidden workflow.}
\label{tab:evolved-rules-audit}
\begin{center}
\small
\setlength{\tabcolsep}{5pt}
\begin{tabular}{lccccc}
\toprule
Backbone & Evolved & Fully correct & Partially correct & Incorrect & High-value \\
\midrule
DeepSeek-V4-Pro & 10 & 8 & 2 & 0 & 9 \\
GLM-5.3-Flash & 15 & 14 & 1 & 0 & 10 \\
GPT-5.6-Terra & 9 & 7 & 2 & 0 & 7 \\
Kimi-K3 & 20 & 20 & 0 & 0 & 10 \\
\bottomrule
\end{tabular}
\end{center}
\end{table}

In Table~\ref{tab:evolved-rules-audit}, \textbf{Evolved} denotes the total
number of verified facts after near-duplicate pruning and merging, where
each fact describes a transition $r:(a,s,c)\longrightarrow\Delta s$.
\textbf{Fully correct} indicates that the mechanism and its critical
conditions match the gold workflow. \textbf{Partially correct} indicates
a valid core mechanism with an overly narrow or broad condition, such as
restricting department-triggered article flagging to drafts or omitting a
manager-reassignment exception in a deactivation cascade.
\textbf{Incorrect} indicates a mechanism contradicted by the environment.
The audit finds no incorrect core mechanisms, while the partially correct
facts identify remaining errors in rule scope. \textbf{High-value} counts
facts describing task-critical hidden workflows, excluding trivial
tool-response echoes, default assignments, and uninformative negative
observations.

The resulting ledgers contain 9--20 facts, of which 7--10 per backbone
describe high-value hidden workflows. Kimi-K3 has the largest ledger and
all 20 facts are fully correct under this audit. DeepSeek-V4-Pro obtains
the highest downstream STCR with only 10 facts: rule count alone does not
determine execution quality, which also depends on how well the backbone
uses the ledger during execution.

\section{Case Study: Hardware-Asset Cost Revaluation}
\label{app:case-cost}

\paragraph{Task.}
The source episode is one World of Workflows training task.  Its description
asks the agent to register six hardware assets for two senior engineers.
Create a PowerEdge T610, tag D001, valued at \$10,500, and assign it to Imani
Okafor.  Register a ThinkStation S20, tag I001, worth \$7,000, and an iPhone
X, tag I002, worth \$6,000, both assigned to Elara Petrov.  Register an iPhone
4, tag A001, valued at \$8,000, for Elara Petrov, and an iPad 3, tag A002,
worth \$9,000, for Imani Okafor.  Register a Laserjet 4240n, tag D002, valued
at \$5,000, for Imani Okafor.  Then transfer D002 from Imani Okafor to Elara
Petrov, A001 from Elara Petrov to Imani Okafor, and A002 from Imani Okafor to
Elara Petrov, and mark Elara Petrov inactive.  The costs are literal amounts.
The text does not mention a markup, a fee, or any later correction.

\paragraph{Evaluator feedback.}
GLM-5.3-Flash created all six assets, received a successful create response
for each, finished the three transfers, deactivated Elara Petrov, and stopped.
None of those create responses contained \texttt{cost}, and the trajectory
never called \texttt{update\_hardware\_asset}.  The agent did not read the
stored costs back and did not correct them.  The evaluator still returned
only one completion error: asset D001 has cost 12575, expected 10500.

That single error is an artifact of the check order.  The evaluator walks
the assets in task order and returns at the first mismatch.  D001, the only
asset submitted at 10500, is first.  The checks for 7000, 6000, 8000, 9000,
and 5000 never run, so their rewritten costs never appear in the feedback.

\paragraph{Goal-judgment hypothesis.}
Goal judgment saw one failed asset and five assets that the evaluator did
not mention.  It treated the silence as evidence that those five costs had
persisted. Its initial hypothesis can be summarized as follows:

\begin{quote}
Creating a hardware asset with a submitted cost above \$10,000 may trigger an
automatic cost revaluation after creation. The task requested a cost of
10500, but the evaluator reported a persisted value of 12575. Assets
submitted at or below 10000 may have kept their submitted costs.
The trigger, timing, and exact adjustment are unknown.
\end{quote}

\paragraph{Abduction.}
The abduction agent tested this hypothesis in a fresh sandbox.  It did not
resume the source episode.  Each step below records the action, the
observation, and the hypothesis the agent kept afterward.  The arithmetic
relation appears only when the agent first wrote it down.

\begin{enumerate}
\item \textbf{Locate the field.}
The agent reads the schema of \texttt{alm\_hardware} and \texttt{alm\_asset}.
\emph{Hypothesis.} Unchanged.  A cost above 10000 may be rewritten after
creation.  The formula is unknown, and a successful create response will not
be treated as proof that the submitted cost persisted.

\item \textbf{Prepare a fresh record.}
The agent creates a new user and resolves a PowerEdge model.
\emph{Hypothesis.} Unchanged.

\item \textbf{Reproduce the source value.}
The agent creates a PowerEdge T610 at 10500 and reads
\texttt{alm\_hardware}.  The stored cost is 12575.  The record was updated by
\texttt{admin} about one second after creation, with \texttt{sys\_mod\_count}
equal to 1.
\emph{Hypothesis.} The 10500 to 12575 change appears to follow creation. Whether
10000 is a threshold, and what the adjustment rule is, are still open.

\item \textbf{Test the threshold.}
The agent creates two more PowerEdge T610 assets, at 9000 and at 10000, and
reads all three hardware records.  The stored costs are 10850 and 12000.
\emph{Hypothesis, revised.} The same rewrite occurs at 9000 and at 10000, so
the agent drops the \$10,000 threshold.  It is at this revision, and not
before, that the agent writes the three pairs as
\begin{equation}
  \mathrm{cost}_{\mathrm{stored}}
  = 1.15 \times \mathrm{cost}_{\mathrm{submitted}} + 500.
  \label{eq:cost-rewrite}
\end{equation}
Whether other models follow the same relation is still open.  The source
episode never reported a cost error on them, because the evaluator had
stopped at D001.

\item \textbf{Test another cost and another model.}
A ThinkStation S20 submitted at 12000 is stored as 14300.  A PowerEdge T610
submitted at 20000 is stored as 23500.  Both agree with
Eq.~(\ref{eq:cost-rewrite}).
\emph{Hypothesis.} The relation is not limited to the original 10500 point or
to a single model. The agent still treats the source episode's apparently unchanged
lower-cost assets as a possible exception.

\item \textbf{Test a later edit.}
The agent calls \texttt{update\_hardware\_asset} and sets the first probe back
to 10500, then reads it again.  The stored cost stays 10500, and
\texttt{sys\_mod\_count} is 2.
\emph{Hypothesis, revised.} Eq.~(\ref{eq:cost-rewrite}) applies to creation.
An explicit cost update after creation is kept as submitted.  The agent also
writes a tentative scope: the source ThinkStation at 7000, and the iPad,
iPhone, and Laserjet assets, may lie outside the rule.

\item \textbf{Look for an implementing script.}
The agent searches \texttt{sys\_script} for a cost rule on
\texttt{alm\_hardware} or \texttt{alm\_asset}.  No such script is returned.
\emph{Hypothesis.} Unchanged.  The relation remains a behavioral rule from the
read-backs.

\item \textbf{Test the suspected exceptions.}
The agent creates an iPad 3 at 9000 and a ThinkStation S20 at 9000.  Both are
stored as 10850.  It also creates a Laserjet 4240n at 5000 and does not read
it yet.
\emph{Hypothesis, revised.} The fresh probes remove the model exception.
Eq.~(\ref{eq:cost-rewrite}) holds for these models at 9000. Whether it holds
below 9000 is still open.

\item \textbf{Test the lower end, then accept.}
The agent creates a ThinkStation S20 at 7000, an iPhone X at 6000, and an
iPhone 4 at 8000, and reads them together with the Laserjet.  The stored
costs are 8550, 7400, 9700, and 6250.  These agree with
Eq.~(\ref{eq:cost-rewrite}).
\emph{Verified rule.} Creating a hardware asset with an explicit cost rewrites
the persisted cost, about one second after insert, by
Eq.~(\ref{eq:cost-rewrite}). On the tested range 5000--20000, the rewrite
does not depend on a 10000 cutoff or on the asset model.  A later
\texttt{update\_hardware\_asset} is exempt.  The rule is written into
$\mathcal{W}_S$ with this action guide: after creation, re-read
\texttt{alm\_hardware.cost}; when the task requires the submitted amount,
restore it with an explicit update and read it back.
\end{enumerate}

\section{Abduction Statistics}
\label{app:abduction-stats}

Table~\ref{tab:abduction-stats} summarizes hidden-rule abduction on the WOW
training tasks. Each launched verification run counts as one candidate.
A directly verified candidate is accepted without revising the hypothesis;
a candidate verified after revision is accepted only after at least one
revision. The abduction agent does not emit an explicit rejection, so that
column is zero. Unresolved candidates never receive a verified rule,
including runs that exhaust the revision budget or remain incomplete.
Average steps is the mean number of actual environment interactions across
all launched runs, including repeated checks and unresolved or incomplete
runs. Only the final rule count is deduplicated; verification outcomes and
interaction costs are not.

\begin{table}[h]
\caption{Abduction outcomes and actual interaction costs across all launched
checks, including one incomplete GLM-5.3-Flash run. Dedup.\ rules counts the
final distinct rules after merging.}
\label{tab:abduction-stats}
\begin{center}
\small
\setlength{\tabcolsep}{4pt}
\begin{tabular}{lccccccc}
\toprule
Backbone & Candidates & Direct & Revised & Rejected & Unresolved & Avg.\ steps & Dedup.\ rules \\
\midrule
DeepSeek-V4-Pro & 14 & 10 & 3 & 0 & 1 & 14.7 & 10 \\
GLM-5.3-Flash & 16 & 13 & 2 & 0 & 1 & 19.1 & 15 \\
GPT-5.6-Terra & 11 & 7 & 2 & 0 & 2 & 13.3 & 9 \\
Kimi-K3 & 23 & 20 & 1 & 0 & 2 & 16.5 & 20 \\
\bottomrule
\end{tabular}
\end{center}
\end{table}

\section{Additional Evolution Round}
\label{app:more-budgets}

Besides the three evolution rounds shared by all methods, \evolutionname{}
spends additional environment interactions on hidden-rule abduction during
evolution. On GLM-5.3-Flash, for example, it launches 16 checks with 19.1
interactions each on average (Appendix~\ref{app:abduction-stats}). To
compare the methods under a closer evolution budget, we give MemoHarness and
WorldEvolver one additional evolution round on GLM-5.3-Flash, providing them
with extra training trajectories and update opportunities.
Table~\ref{tab:more-budgets} shows that their STCR, TCR, and PSR remain
unchanged from the three-round results in Table~\ref{tab:main}, and
\evolutionname{} retains a 10-point STCR advantage over MemoHarness and a
30-point advantage over WorldEvolver. The gap therefore cannot be closed simply
by giving the baselines more evolution: additional trajectory-based
updates do not substitute for targeted verification of hidden rules.

\begin{table}[h]
\caption{Safe task completion rate (STCR), task completion rate (TCR), and policy safety rate (PSR) on GLM-5.3-Flash after one additional evolution round for MemoHarness and WorldEvolver, compared with \evolutionname{}.}
\label{tab:more-budgets}
\begin{center}
\small
\setlength{\tabcolsep}{6pt}
\begin{tabular}{lcccc}
\toprule
Method & Evolution rounds & STCR & TCR & PSR \\
\midrule
MemoHarness & 4 & 65.0 & 75.0 & 90.0 \\
WorldEvolver & 4 & 45.0 & 55.0 & 85.0 \\
\evolutionname{} & 3 & 75.0 & 80.0 & 95.0 \\
\bottomrule
\end{tabular}
\end{center}
\end{table}

\section{Limitations}
The interactive environments make full evolution and evaluation costly,
so we report single-run results for each configuration. 

\section{AI use statement}

We used generative AI tools to polish the language of the manuscript and to assist with writing, as well as to help implement experimental and analysis code.  We have reviewed all AI-assisted text and code, and take full responsibility for the final content of this paper, including its claims, wording, and experimental results.

\end{document}